\documentclass{article} 
\usepackage{iclr2024_conference,times}
\usepackage{graphicx} 
\usepackage{wrapfig}
\usepackage{multirow}
\usepackage{xcolor, colortbl}
\definecolor{baselinecolor}{gray}{.9}
\newcommand{\baseline}[1]{\cellcolor{baselinecolor}{#1}}

\usepackage{amsmath,amsfonts,bm}

\def\eqref#1{equation~\ref{#1}}

\def\1{\bm{1}}

\DeclareMathAlphabet{\mathsfit}{\encodingdefault}{\sfdefault}{m}{sl}
\SetMathAlphabet{\mathsfit}{bold}{\encodingdefault}{\sfdefault}{bx}{n}

\usepackage{hyperref}
\usepackage{url}

\title{Seer: Language Instructed Video Prediction with Latent Diffusion Models}

\author{Xianfan Gu\textsuperscript{\textdagger}\quad  Chuan Wen\textsuperscript{$\ddagger$}\textsuperscript{\textdagger}\textsuperscript{$\diamondsuit$}\quad Weirui Ye \textsuperscript{$\ddagger$}\textsuperscript{\textdagger}\textsuperscript{$\diamondsuit$}\quad Jiaming Song\textsuperscript{$\spadesuit$}\quad Yang Gao\textsuperscript{$\ddagger$}\textsuperscript{\textdagger}\textsuperscript{$\diamondsuit$}\\
\textsuperscript{\textdagger}Shanghai Qi Zhi Institute\quad \textsuperscript{$\ddagger$} IIIS, Tsinghua University\quad \textsuperscript{$\diamondsuit$}Shanghai AI Lab\quad \textsuperscript{$\spadesuit$}NVIDIA\\
\texttt{guxf@sqz.ac.cn}\quad\texttt{\{cwen20,ywr20,gaoyangiiis\}@mails.tsinghua.edu.cn} \\
\texttt{jiamings@nvidia.com}
}

\iclrfinalcopy 
\begin{document}

\maketitle

\begin{abstract}
Imagining the future trajectory is the key for robots to make sound planning and successfully reach their goals. Therefore, text-conditioned video prediction (TVP) is an essential task to facilitate general robot policy learning.
To tackle this task and empower robots with the ability to foresee the future, we propose a sample and computation-efficient model, named \textbf{Seer}, by inflating the pretrained text-to-image (T2I) stable diffusion models along the temporal axis. We enhance the U-Net and language conditioning model by incorporating computation-efficient spatial-temporal attention. Furthermore, we introduce a novel Frame Sequential Text Decomposer module that dissects a sentence's global instruction into temporally aligned sub-instructions, ensuring precise integration into each frame of generation. Our framework allows us to effectively leverage the extensive prior knowledge embedded in pretrained T2I models across the frames. 
With the adaptable-designed architecture, Seer makes it possible to generate high-fidelity, coherent, and instruction-aligned video frames by fine-tuning a few layers on a small amount of data. The experimental results on Something Something V2 (SSv2), Bridgedata and EpicKitchens-100 datasets demonstrate our superior video prediction performance with around 480-GPU hours versus CogVideo with over 12,480-GPU hours: achieving the 31\% FVD improvement compared to the current SOTA model on SSv2 and 83.7\% average preference in the human evaluation.

\end{abstract}

\section{Introduction}

\begin{wrapfigure}[11]{r}{0.5\textwidth}
\vspace{-20pt}
\centering
\includegraphics[width=1.0\linewidth]{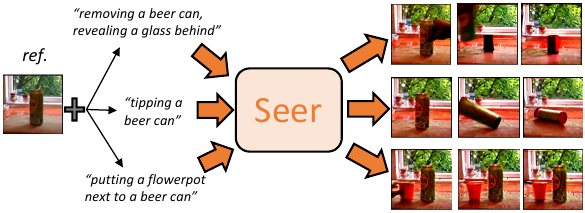}
\vspace{-20pt}
\caption{Seer is an efficient video diffusion model that uses natural language instructions and reference frames (\textit{ref.}) to predict multiple variations of future frames.}
\label{fig:overview}
\end{wrapfigure}
Text-conditioned Video Prediction (TVP), a task that generates future video frames conditioned on a few frames and language instructions, is crucial for diverse downstream tasks requiring instruction alignment and temporal consistency from an initial environment. For instance, in video editing, TVP empowers the generation of diverse temporal movements from an input video clip, guided by a range of language instructions. Importantly, TVP can selectively extend video segments while preserving temporal consistency from the input video. TVP also plays an important role in the scenario of robot learning.  TVP samples coherent future frames with aligned motion trajectories based on the initial state of a robot, providing task-level visual guidance for long-horizon planning. This overcomes the challenge for a robot to align abstract language instructions with long-horizon operations. Consequently, learning a TVP model is a fundamental task to achieve both temporal consistency in the transition distribution of the initial state and alignment between task-level language instructions and video motion, facilitating the development of video foundation models.

\begin{figure*}
\centering
\vspace{-30pt}
\includegraphics[width=0.85\linewidth]{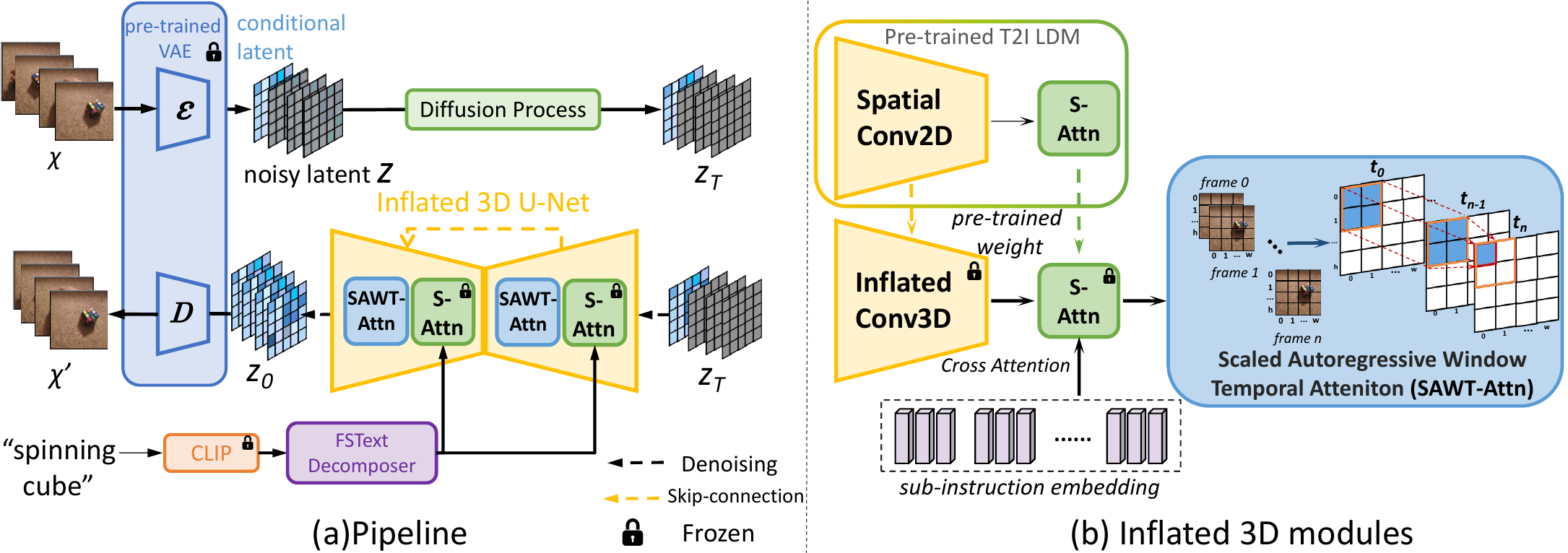}
\vspace{-10pt}
\caption{(a) Seer's pipeline includes an Inflated 3D U-Net for diffusion and a Frame Sequential Text Transformer for text conditioning. (b) Our Inflated 3D U-Net expands the pre-trained 2D Conv kernel to 3D kernels and connects with the Scaled Autoregressive Window Temporal Attention layer.}
\vspace{-15pt}
\label{fig:pipeline}
\end{figure*}
Despite its potential benefits, text-to-video prediction (TVP) is a challenging task because it requires a deep understanding of the initial frames, the natural language instruction, and the grounding between language and images, while predicting based upon all the information above. In contrast to traditional text-to-video generation tasks \citep{godiva,nuwa,hong2023cogvideo,vdm,makeavideo}, which do not explicitly condition on initial frames, TVP requires a model to synthesize predictions based on the given initial frames and textual instructions. Merely generating a few prototypical videos corresponding to the input text is no longer a viable solution in TVP; the task necessitates a more detailed comprehension of temporal movement. 
Besides, the existing text-to-video generation task usually aims to generate short horizon video clips with text specifying the general content, such as “a person is skiing”, while our aim in the TVP task is to use the text as a task descriptor, such as “tipping a beer can”, as shown in Figure~\ref{fig:overview}.  

Specifically, there are mainly three problems limiting the performance of the TVP task: 
\textbf{1) Requirement for large-scale labeled text-video datasets and expensive computational cost:} learning to capture the correspondence between two different modalities is non-trivial and needs large amounts of supervised text-video pairs and excessive computation overhead for training. 
\textbf{2) Low fidelity of generated frames:} the frames generated by the models are usually blurry and cannot clearly display the background and objects specified in the reference frames.
\textbf{3) Lack of fine-grained instruction for each frame in the task-level videos:} the goals specified by text instructions are usually in the task level, making it difficult to understand the progress and generate the corresponding frame in each timestep only conditioned on a global text embedding.
To address these issues, we propose \textbf{Seer}: a TVP method capable of generating task-level videos according to the text guidance with high data and training efficiency.

Motivated by the recent progress on generative models~\citep{ldm,ramesh2022dalle2}, we propose to leverage text-to-image (T2I) latent diffusion models~\citep{ldm} for the TVP tasks. T2I models are pretrained with billions of text-image pairs crawled from the internet~\citep{schuhmann2021laion}. They have acquired rich prior knowledge and thus are able to generate high-quality images corresponding to the text descriptions. Therefore, inheriting such prior knowledge by inflating a T2I model along the temporal axis and fine-tuning it with a small text-video dataset is an appealing solution for TVP tasks, which relieves the requirement for extensive labeled data and computational overhead, i.e., Problem 1.

Since the T2I models contain two modalities: image and language, we propose to inflate these two parts to generate high-quality video frames and fine-grained text instruction embeddings for each timestep respectively. For the visual model, we extend the 2D latent diffusion model~\citep{ldm} to data and computation-efficient 3D network to model spatial dependencies and the temporal dynamics simultaneously, which is called Inflated 3D U-Net.
By taking advantage of joint modeling of spatial and temporal dimensions, as well as autoregressive generation, we successfully synthesize coherent and high-fidelity frames, which alleviates Problem 2.
As for the language module, in contrast to existing approaches~\citep{magicvideo,tuneavideo} that encode one text embedding for the whole video with a text encoder, we propose to decompose the single text instruction into fine-grained guidance embeddings for each time step. We achieve the automatic decomposition by a \textbf{Frame Sequential Text} (FSText) Decomposer based on the causal attention mechanism. By temporally splitting the instruction into different phases, 
Seer improves the guidance embeddings for each frame and thus enables task-level video generation (Problem 3).

We conduct extensive experiments on Something-Something V2~\citep{sthv2},  Bridge Data~\citep{bridge} and Epic-Kitchens-100~\citep{epickitchen} datasets. 
We outperform all the baselines, such as MCVD~\citep{mcvd}, TATS~\citep{tats} and \textit{Tune-A-Video} (TAV)~\citep{tuneavideo}, and achieve state-of-the-art performance in terms of FVD and KVD. 
Especially we improve FVD from $163$ to $113$ on Something-Something V2, compared to the SOTA TAV.
Compare to over 480 hours with $13 \times 8$ A100 GPUs in CogVideo~\citep{hong2023cogvideo}, the experiments show the high efficiency of our method: 120 hours with 4 RTX 3090 GPUs. 
The ablation studies illustrate the effectiveness of our computation-efficient video-inflated model with the newly proposed FSText Decomposer. Furthermore, our method supports video manipulation by modifying the text instructions, and we demonstrate our superior generation quality through a human evaluation study, showing more than 70\% preference over TAV and around 90\% preference over TATS and MCVD.

\section{Related Work}

\subsection{Text-to-Image Generation}
Since Scott Reed et al.~\citep{reed2016generative} firstly set up the T2I generation task and proposed a GAN-based method, this multi-modal generation task has attracted the attention of the computer vision community. 
DALL-E~\citep{ramesh2021dalle} makes a breakthrough by modeling the T2I generation task as a sequence-to-sequence translation task with a VQ-VAE~\citep{van2017vqvae} and Transformer~\citep{vaswani2017attention}. Since then, many variants have been proposed with an improved image tokenizer~\citep{yu2022parti}, hierarchical Transformers~\citep{ding2022cogview2} or domain-specific knowledge~\citep{gafni2022make}.
With the recent progress of Denoising Diffusion Probabilistic Models (DDPM)~\citep{ddpm}, the diffusion models have been widely used for T2I generation tasks~\citep{glide,imagen,ramesh2022dalle2}. Specifically, GLIDE~\citep{glide} proposes classifier-free guidance for T2I diffusion models to improve image quality. For a better alignment between text and image, DALL-E 2~\citep{ramesh2022dalle2} proposed to denoise CLIP~\citep{radford2021clip} image embedding conditioned on CLIP text embedding, which integrated high-level semantic information. To reduce the computation cost of the denoising process in pixel space, Latent Diffusion Model (LDM) employs VAE~\citep{Kingma2014vae} to operate in the latent space. 
Seer takes advantage of the prior language-vision knowledge of pretrained LDM and inflates it along the time axis.

\subsection{Text-to-Video Generation}
In contrast to the huge success of Text-to-Image (T2I) generation, Text-to-Video (T2V) generation is still underexplored due to the limitation of the large text-video data annotation and computing resources.  Inspired by the various variants of T2I generation, recent T2V studies have attempted to explore compatible variants for video generation modeling. GODIVA~\citep{godiva} first proposes a VQ-VAE based auto-regressive model with three-dimensional sparse attention for T2V generation. N\"UWA~\citep{nuwa} further improves it by designing a 3D encoder with 3D nearby attention and achieves competitive performance on multi-task generation. Unlike the single frame-rate T2V approaches trained from scratch on large-scale text-video datasets, CogVideo~\citep{hong2023cogvideo} proposes a multi-frame-rate hierarchical model for T2V generation. This approach leverages the pre-trained module of T2I CogView-2~\citep{ding2022cogview2}.

Motivated by the remarkable progress of T2I diffusion models~\citep{ramesh2022dalle2,imagen,ldm}, Make-A-Video~\citep{makeavideo}, MagicVideo~\citep{magicvideo}, Tune-A-Video~\citep{tuneavideo} and Imagen Video~\citep{imagenvideo} transfer the 2D diffusion models to 3D models by incorporating temporal modules in T2V generation. In contrast to Imagen Video, all other three methods utilize the prior knowledge of T2I pre-trained model. Similarly, we use the pre-trained weight of the 2D T2I diffusion model in our 3D T2V model. Varying from the aforementioned methods, our method Seer utilizes autoregressive attention on both spatial and temporal spaces to generate high-fidelity and coherent video frames. And Seer is able to handle the task-level video prediction by decomposing the language condition into fine-grained sub-instruction.

\section{Preliminaries}
\noindent\textbf{Denoising Diffusion Probabilistic Models with classifier-free guidance:}
Diffusion models are probabilistic models that approximate the data distribution by iteratively adding noise and denoising through a forward/reverse Gaussian Diffusion Process~\citep{ddpm,song2020score}. The forward process applies noise at each time step $t\in{0,...,T}$ to the data distribution $\mathbf{x}_{0}$, creating a noisy sample $\mathbf{x}_t$ where $\mathbf{x}_t = \sqrt{\bar{\alpha}_t}\mathbf{x}_0+\sqrt{1-\bar{\alpha}_t}\bm{\epsilon}$ ($\bm{\epsilon}\sim\mathcal{N}(\boldsymbol{0},\boldsymbol{I})$), and $\bar{\alpha}_t$ is the accumulation of the noise schedule $\alpha_{0:T}$ defined by $\bar{\alpha}_t=\prod^t_{s=1}\alpha_s$. To denoise images, the diffusion process uses a reparameterized variant of Gaussian noise prediction $\bm{\epsilon}_\theta(\mathbf{x}_t,t)$ targeting Gaussian noise $\bm{\epsilon}$. The reverse process $p(\mathbf{x}_{t-1}|\mathbf{x}_{t})$ of the Markov Chain generates new samples from Gaussian noise, which is approximated by Bayes' theorem as $q(\mathbf{x}_{t-1}|\mathbf{x}_t,\mathbf{x}_0)$, where $\mathbf{x}_0$ is derived from the forward process as $\mathbf{x}_0 = \frac{1}{\sqrt{\bar{\alpha}_t}}(\mathbf{x}_t-\sqrt{1-\bar{\alpha}_t\bm{\epsilon}_\theta(\mathbf{x}_t,t)})$.

Classifier-free guidance~\citep{clsfree} is introduced for conditional diffusion models to generate images without requiring an extra image classifier. A conditional model with a parameterized reverse process $p(\mathbf{x}_{t-1}|\mathbf{x}_t,\mathbf{c})$ uses a conditional identifier $\mathbf{c}$ through $\bm{\epsilon}_{\theta}(\mathbf{x}_t,t,\mathbf{c})$. To predict an unconditional score, the conditional identifier is replaced with a null token $\O$ and denoted as $\bm{\epsilon}_{\theta}(\mathbf{x}_t,t,\mathbf{c}=\O)$. Classifier-free guidance can then be approximated as a linear combination of conditional and unconditional predictions:
\vspace{-3pt}
\begin{equation}
   \bm{\tilde{\epsilon}}_{\theta}(\mathbf{x}_t,t,\mathbf{c}) = (1+w)\bm{\epsilon}_{\theta}(\mathbf{x}_t,t,\mathbf{c})-w\bm{\epsilon}_{\theta}(\mathbf{x}_t,t,\mathbf{c}=\O),
   \vspace{-3pt}
\end{equation}
where $w$ is the guidance scale. Text-video and text-image-based diffusion models~\citep{ldm,imagen,glide,vdm,makeavideo} use DDPM with classifier-free guidance. This diffusion method can be adapted to various tasks with flexibility.

\noindent\textbf{Latent Diffusion Models:} 
Compared with image diffusion, video diffusion has significantly higher computation costs because it needs to process multiple frames.
Recent works have explored the computation-efficient version of diffusion modeling, such as latent diffusion model (LDM)~\citep{ldm}. LDM proposes the VAE-based latent diffusion, including a KL-regularized autoencoder for encoding/decoding latent representation $\bm{\varepsilon}(\mathbf{x})$, and a diffusion model to operate on the latent space $\mathbf{z}_t$.
For the conditional generation, LDM introduces a domain-specific encoder $\bm{\tau}_\theta$ to the projection of condition $\mathbf{y}$ for various modality generations. Thus, the objective of LDM is: 
\vspace{-5pt}
\begin{equation}
    \vspace{-10pt}
    L_{\mathrm{LDM}} = \mathbb{E}_{t,\bm{\varepsilon}(\mathbf{x}),\mathbf{y},\bm{\epsilon}\sim\mathcal{N}(\boldsymbol{0},\boldsymbol{I})}\Bigr[\|\bm{\epsilon} - \bm{\epsilon}_\theta(\bm{z}_t,t,\bm{\tau}_\theta(\mathbf{y}))\|^2\Bigr]
\end{equation}

\section{Methodology}\label{sec:method}
In this paper, we aim to explore an efficient diffusion method to predict coherent video frames guided by language instructions, which requires learning to parse natural language, understand the scene, and ground the language and scene together. However, it is challenging to directly apply conventional video diffusion models for TVP due to the following problems: (1) The limited labeled text-video data and computational resources. (2) Low fidelity of frame generation. (3) Lack of fine-grained instruction for each frame in the task-level videos.
 

\subsection{Overview of Seer}
\label{sec:inflate}
Motivated by the robust generative capabilities of text-to-image (T2I) diffusion models, we leverage the prior knowledge implied in pretrained T2I models by inflating the 2D U-Net~\citep{ldm} and incorporating temporally consistent layers. However, the inflated video diffusion model guided solely by coarse global language instruction tends to generate irrelevant T2I outcomes and fails to maintain temporal coherency between video frames. To address this limitation and provide precise and controllable guidance for our inflated model, we introduce a novel temporal decomposition component for language instruction, this component decomposes global instruction as temporally aligned sub-instruction for delicate task-level guidance, which significantly enhances the fidelity and coherency of predicted video.

Our Seer method comprises two main components: the video diffusion and the language conditioning modules. We propose to enhance these two components to facilitate high-fidelity frame synthesis and the temporal alignment of text instructions, respectively. Specifically, as shown in Figure~\ref{fig:pipeline} (a), we utilize two pathways to implement the conditional diffusion process guided by reference frames and language: \textbf{1)} We incorporate the spatial-temporal module discussed in Section~\ref{sec:efficientnet} into the Inflated 3D U-Net. This integration enables the propagation of contextual information from reference frames to future frames within the spatial-temporal space, allowing for coherent motion prediction based on the reference frames. \textbf{2)} To plan continuous motion with fine-grained language guidance, we introduce a Frame Sequential Text (FSText) Decomposer in Section~\ref{sec:fstext}. This module transforms global language instructions into multi-timestep sub-instructions that are synchronized with video. Subsequently, we inject these frame-wise subinstruction tokens into the intermediate latent space of the video frames at each time step.
With this design, we merely train the spatial-temporal layers and FSText module from scratch while freezing the remaining pretrained modules within our 3D inflated U-Net. These two modules are jointly trained by the diffusion objective, where $f_\theta$ is our FSText decomposer, $\bm{\tau}$ is the frozen CLIP text encoder, and $\mathbf{y}$ is the input text:
\begin{equation}
    L_{\mathrm{diffusion}} = \mathbb{E}_{t,\bm{\varepsilon}(\mathbf{x}),\mathbf{y},\bm{\epsilon}\sim\mathcal{N}(\boldsymbol{0},\boldsymbol{I})}\Bigr[\|\bm{\epsilon} - \bm{\epsilon}_\theta(\mathbf{z}_t,t,f_\theta(\bm{\tau}(\mathbf{y})))\|^2\Bigr],
\end{equation}


\subsection{Data \& Computation-efficient 3D Network}\label{sec:efficientnet}
To design a computation-efficient visual backbone for our video diffusion model,  we refer to some relevant works on lifting 2D to 3D video modeling~\citep{i3d} and efficient attention computation~\citep{swin, videoswin}. In general, we leverage the latent diffusion model (LDMs) pretrained on T2I tasks to build a text-video model. Our inflated 3D U-Net consists of two principal components as illustrated in Figure~\ref{fig:pipeline} (b): \textbf{1)} The 3D spatial layers, where we draw inspiration from I3D~\citep{i3d} and enhance the 2D convolution kernel from ($3 \times 3$) to a 3D counterpart ($1 \times 3 \times 3$)  with an added video frames axis from the pre-trained 2D modules, consisting of a series of 2D ResNet blocks and Spatial Attention Blocks.
\begin{wrapfigure}[17]{r}{0.42\textwidth}
\centering
\vspace{-10pt}
\includegraphics[width=0.8\linewidth]{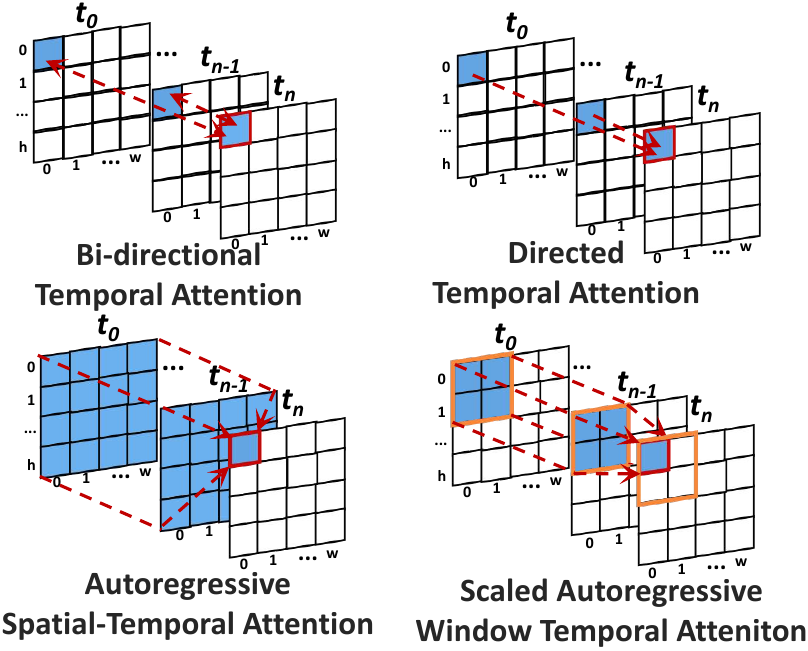}
\vspace{-10pt}
\caption{Variants of temporal attention, only the blue tokens attend to the current token in the red box. Red dashed arrows indicate the direction of attention. And the orange boxes indicate the local window region ($2\times 2$ window in this case).}
\label{fig:tempattn}
\end{wrapfigure}
\textbf{2)} The temporal layers, play a crucial role in our visual backbone for propagating contextual information from the reference frame's image prior across the temporal sequence. We investigated various temporal attention and incorporated them into our 3D U-Net architecture. Our empirical observations indicate that bi-directional temporal attention tends to disregard guidance from reference frames, and both bi-directional and directed temporal attention struggle to capture dependencies among spatial regions, as discussed in Section~\ref{sec:ablate:temp}. To address these limitations while reducing complexity, we employ an efficient approach that builds upon the concept of window attention~\citep{swin} in 3D space: the implementation of local window attention in an autoregressive manner across spatial-temporal dimensions. As illustrated in Figure~\ref{fig:tempattn}, we establish fixed local windows for each spatial region with a window size of $m \times m$ relative to the global frame sequence $n$. Within this framework, we compute self-attention using a causal mask, considering both local spatial and global temporal dimensions within the 3D space. This effectively constrains pixel propagation from the future temporal-spatial sequence.

Finally, We maintain the acquired knowledge from the 2D modules by freezing all pretrained weights and exclusively training the spatiotemporal attention layers during fine-tuning. Overall, through a combination of frozen pre-trained spatial layers and lightweight spatiotemporal layers, our inflated 3D U-Net not only retains crucial knowledge but also enhances fine-tuning efficiency.

\subsection{Frame Sequential Text Decomposer}\label{sec:fstext}
For the language conditioning module, since our 3D inflated U-Net is built upon a pretrained text-to-image model, we noticed that using a text-to-image prior alongside a global instruction tends to provide strong semantic guidance, which can override the scene in reference frames, deviating from the intended guidance for prediction based on the existing scenes.
To address the above limitation and better capture long-term dependencies from both text and reference frames, we introduce the Frame Sequential Text (FSText) Decomposer. This novel approach decomposes the global instruction into fine-grained sub-instructions, aligning with each frame. We further explore the interpretability of sub-instruction embeddings in Section~\ref{sec:results:subins}.
\begin{figure*}
\centering
\vspace{-30pt}
\includegraphics[width=0.9\linewidth]{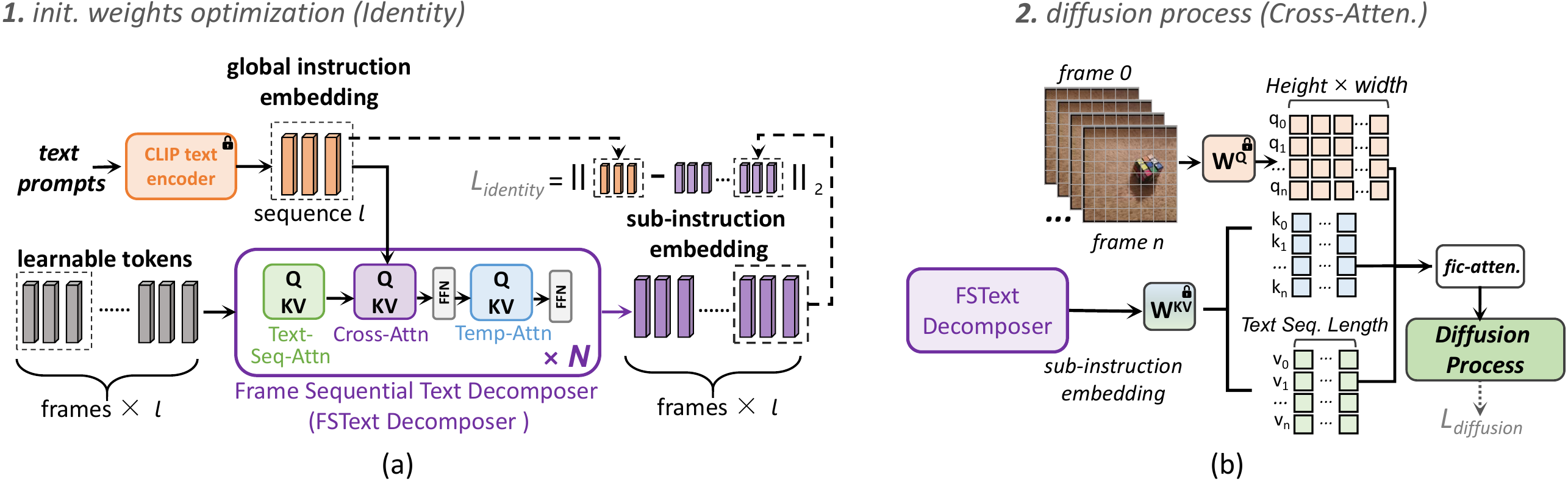}
\vspace{-10pt}
\caption{Frame Sequential Text Decomposer is shown in (a). We start by initializing the weight of the network to project identity vectors from CLIP text tokens. We then optimize the generated text tokens via the diffusion process (b), where frame-individual cross-attention is denoted by ``fic-attn."}
\label{fig:fseq}
\vspace{-10pt}
\end{figure*}
To derive a sequence of temporally aligned sub-instruction embeddings from the global instruction generated by the CLIP text encoder~\cite{radford2021clip}, we employ a transformer-based temporal network designed to fulfill three essential properties for meaningful sub-instructions: \textbf{1)} Contextual aggregation, which ensures that the inner tokens of each sub-instruction aggregate contextual information within the sentence. \textbf{2)} Semantic inheritance, the semantic information of these sub-instructions is inherited directly from the global instruction \textbf{3)} Temporal consistency ensures alignment between the sub-instructions and the time sequence, thereby facilitating the generation of temporally consistent video. Based on these properties,  our network consists of  three key components: \textbf{a)} To achieve the property of contextual aggregation, we employ Text-Sequential Attention, akin to BERT, a bidirectional self-attention layer~\citep{bert} to capture global dependencies among different positions within text sentences. \textbf{b)} To ensure semantic inheritance, we use Cross-Attention, responsible for projecting the global instruction's textual sequence onto the inner tokens of each sub-instruction, this component ensures that all sub-instructions contain essential global instruction signals for guiding video frame generation. \textbf{c)} To maintain temporal consistency, we adopt temporal Attention, a directed attention layer to capture temporal dependencies along the frame axis, which enhances temporal consistency among the generated sub-instructions throughout the video.
\begin{wrapfigure}[10]{r}{0.4\textwidth}
\centering
\vspace{-10pt}
\includegraphics[width=0.9\linewidth]{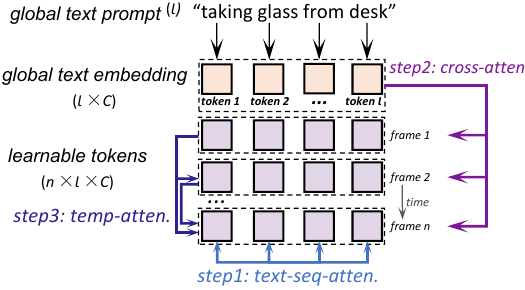}
\vspace{-10pt}
\caption{The FSText attention of sub-instruction tokens.}
\label{fig:fstextpipline}
\end{wrapfigure}
Specifically, as shown in Figure~\ref{fig:fstextpipline}, we start with a global CLIP text embedding, denoted as $(l, C)$, where $l$ signifies the text sentence length and $C$ is the channel size, we initialize learnable tokens with shape $(n, l, C)$ where $n$ denotes the number of frames. The tokens are fed into the text sequential attention layer to perform self-attention along the $l$ axis. Subsequently, the cross-attention layer employs these learnable tokens as queries and the global text embedding as keys and values, resulting in a one-to-multiple projection from the global text into $n$ time steps. This yields $(n, l, C)$ tokens for $n$ frame containing task instruction information. Finally, the temporal attention layer conducts directed attention along the $n$ axis for each token in the textual sequence, transforming the macro-instruction progress into frame-specific guidance.

After getting $n$ sub-instruction embeddings corresponding to each frame, the next step is to inject this guidance into the diffusion process, which is commonly completed by a cross-attention layer. As shown in Figure~\ref{fig:fseq} (b), different from the existing works~\citep{magicvideo,tuneavideo} that calculate the cross-attention between the global instruction embedding and $n$ frames. In our cross-attention layer, where cross-attention is calculated separately between visual latent vectors and sub-instruction embeddings for each frame, and the results from all frames are then concatenated, an attention mechanism we refer to as frame-individual cross-attention (\textit{fic-attn}).

\paragraph{Initialization~~} We find initialization is critical to FSText decomposer. Especially, the random initialization fails to approximate the distribution of text embeddings in the pretrained T2I model and results in poor performance. To guarantee the sub-instruction embeddings become a close approximation of the CLIP text embedding, we employ an initialization strategy by enforcing the FSText decomposer to be an identity function (Note that this initialization step is completed before the diffusion process. We ablate this design in Section~\ref{sec:ablate:fstext}). It can be achieved by this objective:
\begin{equation}
    L_{\mathrm{identity}} = \|f_\theta(\bm{\tau}(\mathbf{y})) - \bm{\tau}(\mathbf{y})\|^2
\end{equation}

\section{Experiments}
In this section, we evaluate Seer on the text-conditioned video prediction task. 
We compare against various recent methods and conduct ablation studies on the techniques presented in Section~\ref{sec:method}.

\subsection{Datasets}\label{sec:dataset}
We conduct experiments on three text-video datasets: Something Something-V2 (SSv2)~\citep{sthv2}, which contains videos of human daily behaviors with language instructions, BridgeData~\citep{bridge} that is rendered by a photo-realistic kitchen simulator with text prompts, and EpicKitchens-100~\citep{epickitchen} (Epic100), which collects human daily activities in the kitchen in egocentric vision with multi-language narrations. For SSv2, we follow \citep{ucf101} to evaluate the first 2048 samples during evaluation to save testing time.
For BridgeData, we split the dataset into an $80\%$ training set and $20\%$ validation set for evaluation. To reduce complexity, we downsample each video clip to 12 frames for SSv2 and Epic100, and 16 frames for BridgeData during training/evaluation. Besides, We provide the zero-shot evaluation on EGO4D~\citep{ego4d} dataset in section~\ref{sec:generlize}. Moreover, we also included an additional evaluation on the UCF-101 dataset~\citep{ucf101} in Appendix~\ref{appendix:sec:ucf101}.

\subsection{Implementation Details}\label{sec:impl}
\begin{wrapfigure}[14]{r}{0.45\textwidth}
\centering
\vspace{-35pt}
\includegraphics[width=1.0\linewidth]{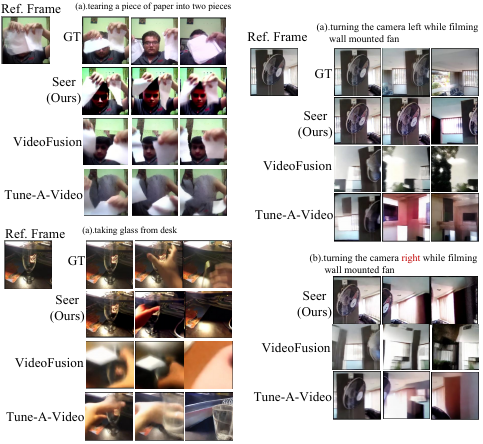}
\vspace{-20pt}
\caption{Visualization of Text-conditioned Video Prediction with the original (a) and manually modified (b) text prompts on SSv2.}
\label{fig:tvm:sthv2}
\end{wrapfigure}
We use the pre-trained weights of Stable Diffusion-v1.5~\citep{ldm} to initialize the VAE, ResNet Blocks and Spatial-Cross Attention layers of the 3D U-Net. We freeze both the pre-trained VAE, and only fine-tune the SAWT-Atten layers in our 3D U-Net. To fine-tune the FSText Decomposer, we initialized it as the identity function of the CLIP text embedding, as described in Section~\ref{sec:fstext}. We train the models with an image resolution of $256 \times 256$ on Something Something-V2 for 200k training steps, EpicKitchens-100 and BridgeData for 80k training steps. In the evaluation stage, we speed up the sampling process with the fast sampler DDIM~\citep{ddim} and conditional guidance of 7.5 for 30 timesteps. See more details in Appendix~\ref{appendix:sec:impl}.

\subsection{Evaluation Settings}
\noindent\textbf{Baselines.~~}\label{sec:baseline} We compare Seer with seven publicly released baselines for video generation (1) conditional video diffusion methods: \textit{Tune-A-Video}~\citep{tuneavideo}, \textit{Masked Conditional Video Diffusion} (MCVD)\citep{mcvd}, \textit{Video Probabilistic Diffusion Models} (PVDM)\citep{pvdm} and VideoFusion\citep{videofusion}; (2) autoregressive-based transformer method: \textit{Time-Agnostic VQGAN and Time-Sensitive Transformer} (TATS)\citep{tats} and \textit{Make It Move} (MAGE)\citep{mage2}; and (3) CNN-based encoder-decoder: SimVP\citep{simvp}.

\noindent\textbf{Machine Evaluation.~~}\label{sec:exp:tvp} We evaluate the text-conditioned video prediction of several baseline methods on Something Something-V2 (SSv2) (with 2 reference frames), Bridgedata (with 1 reference frame) and Epic-Kitchens-100 (Epic100) (with 1 reference frame). Additionally, we conduct several ablation studies of our proposed modules on SSv2. We report the Fréchet Video Distance (FVD) and Kernel Video Distance (KVD) metrics in our evaluation. FVD and KVD are calculated with the Kinetics-400 pre-trained I3D model~\citep{i3d}. We evaluate FVD and KVD on 2,048 SSv2 samples, 5,558 Bridgedata samples and 9,342 Epic100 samples in the validation sets. For FVD metrics, we follow the evaluation code of VideoGPT~\citep{videogpt}. We further evaluate the class-conditioned video prediction of our method on the UCF-101 dataset and present the comparison results in Appendix~\ref{appendix:sec:ucf101}.

\begin{wrapfigure}[7]{r}{0.5\textwidth}
\centering
\vspace{-15pt}
\includegraphics[width=0.8\linewidth]{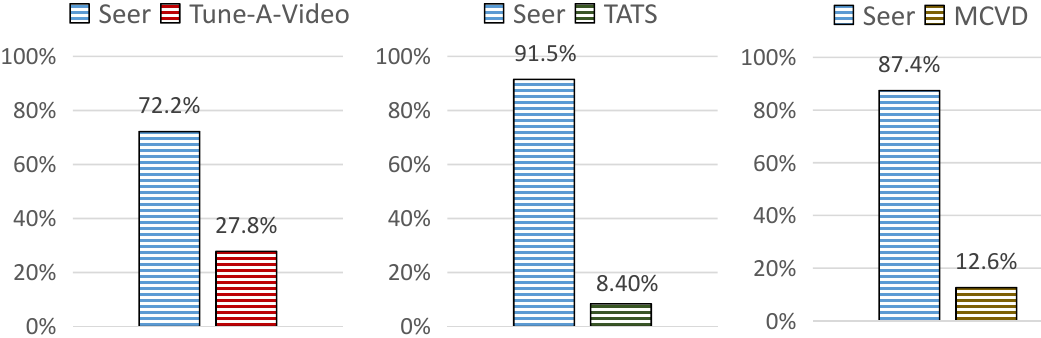}
\vspace{-10pt}
\caption{Human evaluation results. Preference percentage for TVM task on SSv2.}
\label{fig:humaneval}
\end{wrapfigure}

\noindent\textbf{Human Evaluation.~~}\label{sec:exp:humaneval} Besides evaluating the models on the standard validation sets, we also manually modify the text prompts to provide richer testing results, called text-conditioned video manipulation. Because of the absence of ground-truth frames, we conducted a human evaluation of text-conditioned video manipulation (TVM) using 99 video clips from the validation set of SSv2. We manually modified partial text prompts and generated 99 predicted videos for each method. Then, we invited 54 anonymous evaluators to rate the quality of the prediction, with a higher priority placed on the semantic contents in the videos and an intermediate priority placed on the fidelity of the video frames. We report the overall preference choices among the 99 video clips. More details are introduced in Appendix~\ref{appendix:sec:humaneval}

\subsection{Main Results}\label{sec:main-results}
\noindent\textbf{Quantitative Results.~~} Table~\ref{table:tvp} presents the text-condtioned video prediction results on Something Something-V2 (SSv2),  BridgeData and Epic-kitchens-100 (Epic100). Seer achieves the best performance among all baselines, with the lowest Fréchet Video Distance (FVD) of 112.9 and Kinematic Distance (KVD) of 0.12 in SSv2, the lowest FVD of 246.3 and KVD of 0.55 in BridgeData, and the lowest FVD of 271.4 in Epic100. Notably, Seer, MAGE, VideoFusion and Tune-A-Video all incorporate text conditioning, and the results highlight Seer's superior text-video alignment performance.

The results of the human evaluation in the text-conditioned video manipulation experiment are shown in Figure~\ref{fig:humaneval}. Our proposed Seer outperforms the other baselines in terms of both semantic content and fidelity of video, with a preference rate of at least $72.2\%$ in comparison. This indicates that Seer is effective in generating high-quality video clips that are faithful to the input text prompts.

\begin{wrapfigure}[10]{r}{0.47\textwidth}
\centering
\vspace{-10pt}
\includegraphics[width=0.9\linewidth]{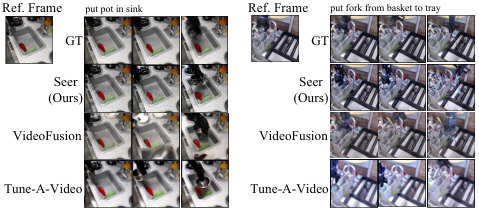}
\vspace{-10pt}
\caption{Visualization of Text-conditioned Video Prediction on Bridgedata.}
\label{fig:tvp:bridge}
\end{wrapfigure}
\noindent\textbf{Qualitative Results.~~} Figure~\ref{fig:tvm:sthv2} compares the text-conditioned video prediction and manipulation performance of Seer, VideoFusion and Tune-A-Video on Something Something-V2 (SSv2). Seer performs better in handling the temporal dynamics of the video and achieving more precise text-video alignment in video manipulation. 
For instance, consider the task of ``turning the camera left" Seer seamlessly generates coherent movement while preserving the background during the camera view adjustment.  In contrast, both VideoFusion and Tune-A-Video exhibit semantic movement but fail to maintain temporal consistency in the video background. Additionally, Seer can imagine hidden objects by utilizing its text-to-image diffusion prior. This flexibility allows Seer to effectively address occlusion in video prediction. In the ``tearing paper" sample, Seer accurately predicts that a man is hidden behind the paper and generates coherent frames including the man's face. Figure~\ref{fig:tvp:bridge} compares Seer, VideoFusion and Tune-A-Video's TVP performance on Bridgedata, illustrating that Seer achieves better text-video alignment of instructed behavior and target objects in future frames, and predicts a more coherent video with higher fidelity.

\begin{table*}
\centering\small
\vspace{-20pt}
\setlength{\tabcolsep}{1.5pt}
{\caption{\textbf{ Text-conditioned video prediction (TVP) results on Something-Something V2 (SSv2), Bridgedata (Bridge), and Epic-Kitchens-100 (Epic100).} We report the FVD and KVD metrics of each method in SSv2,  Bridgedata, and Epic100.
}
\label{table:tvp}}
\begin{tabular}{cccc|cc|cc|cc}
 \multirow{2}{*}{Method} & \multirow{2}{*}{Pre.-weight} & \multirow{2}{*}{Text} & \multirow{2}{*}{Resolution} & \multicolumn{2}{c|}{\textbf{SSv2}} & \multicolumn{2}{c}{\textbf{Bridge}} & \multicolumn{2}{c}{\textbf{Epic100}}\\
   &  &  &  & FVD$\downarrow$ & KVD$\downarrow$  & FVD$\downarrow$ & KVD$\downarrow$  & FVD$\downarrow$ & KVD$\downarrow$\\
 \hline
TATS~\citep{tats} & video & No  & $128\times 128$ & 428.1 & 2177 & 1253 & 6213 & 920.0 & 5065\\
 MCVD~\citep{mcvd} & No & No  & $256\times 256$ & 1407 & 3.80 & 1427 & 2.50 & 4804 & 5.17\\
 SimVP~\citep{simvp} & No & No  & $64\times 64$ & 537.2 & 0.61 & 681.6 & 0.73 & 1991 & 1.34\\
MAGE~\citep{mage2} & video & Yes  & $128\times 128$ & 1201.8 & 1.64 & 2605 & 3.19 & 1358 & 1.61\\
PVDM~\citep{pvdm} & No & No & $256\times 256$ & 502.4 & 61.08 & 490.4 & 122.4 & 482.3 & 104.8\\
VideoFusion~\citep{videofusion} & txt-video & Yes & $256\times 256$ & 163.2 & 0.20 & 501.2 & 1.45 & 349.9 & 1.79\\
Tune-A-Video~\citep{tuneavideo} & txt-img & Yes & $256\times 256$ & 291.4 & 0.91 & 515.7 & 2.01 & 365.0 & 1.98\\
Seer (Ours) & txt-img & Yes & $256\times 256$ & $\bf 112.9$ & $\bf 0.12$ & $\bf 246.3$ & $\bf 0.55$ & $\bf 271.4$ & $ 1.40$\\
\end{tabular}
\vspace{-10pt}
\end{table*}

\subsection{Ablation study}
In this section, we evaluate the effect of different components of our method in the TVP task on the SSv2 dataset. We also evaluate the zero-shot TVP task of different models on the EGO4D dataset.

\begin{wraptable}[7]{r}{0.32\textwidth}
\vspace{-20pt}
\centering\small
\setlength{\tabcolsep}{4pt}
\caption[temp]{\textbf{Ablation study of temporal attention}}
\begin{tabular}{c|cc}
temp. attn. & FVD$\downarrow$ & KVD$\downarrow$\\
 \hline
 \textit{bi-direct.} & 258.2 & 0.56\\
 \textit{directed.} & 222.3 & 0.40\\
 \textit{autoreg.} &200.1 & 0.30\\
 \textit{win-auto.}(Ours) &112.9 & 0.12\\
\end{tabular}
\label{table:ablation:tempattn}
\end{wraptable}
\noindent\textbf{Temporal Attention.~~}\label{sec:ablate:temp}
As shown in Table~\ref{table:ablation:tempattn} studies the effectiveness of different types of temporal attention. Our scaled autoregressive window temporal attention (win-auto.) outperforms autoregressive spatial-temporal attention (autoreg.), bi-directional temporal attention (bi-direct.) and directed temporal attention (directed.), resulting in the lowest FVD and KVD scores. We also find that directed temporal attention further improves video prediction performance compared to bi-directional temporal attention because it utilizes the inductive bias of sequential generation.

\begin{wraptable}[5]{r}{0.32\textwidth}
\vspace{-20pt}
\centering\small
\setlength{\tabcolsep}{4pt}
\caption[ftext]{\textbf{Init. weight ablation results of FSText}}
\begin{tabular}{c|cc}
init. weight & FVD$\downarrow$ & KVD$\downarrow$\\
 \hline
 \textit{random} & 367.9 & 0.75\\
\textit{identity}(Ours) & 112.9 & 0.12\\
\end{tabular}
\label{table:ablation:weight}
\end{wraptable}

\noindent\textbf{FSText Decomposer.~~}\label{sec:ablate:fstext}
Table~\ref{table:ablation:weight} compares different weight initialization strategies of FSText decomposer. The results show that using identity initialization described in Section~\ref{sec:fstext} yields higher prediction quality compared with random initialization.  
This finding demonstrates that identity initialization is necessary for the temporal-text projection of FSText decomposer. See additional ablation results in Appendix~\ref{appendix:sec:fstext}.

\begin{wraptable}[6]{r}{0.47\textwidth}
\vspace{-20pt}
\centering\small
\setlength{\tabcolsep}{4pt}
\caption[fine]{\textbf{Ablation study of Fine-tune settings}}
\begin{tabular}{cc|cc} 
fine-tune& FSText. & FVD$\downarrow$ & KVD$\downarrow$\\
 \hline
 \textit{temp-attn.} & & 328.2 & 1.26\\
 \textit{cross}+\textit{temp-attn.} & & 249.9 & 0.73\\
 \textit{temp-attn.}(Ours) & \checkmark &112.9 & 0.12\\
 \textit{cross}+\textit{temp-attn.} & \checkmark & 1807 & 5.12\\
\end{tabular}
\label{table:ablation:finetune}
\end{wraptable}
\noindent\textbf{Fine-tune Setting.~~}
We compare various fine-tuning settings of 3D Inflated U-Net in Table~\ref{table:ablation:finetune}. Our default setting involves fine-tuning both FSText decomposer (FSText.) and scaled autoregressive window temporal attention (SAWT-Attn.) layers (\textit{temp-attn.}), while freezing the remaining modules in 3D U-Net. For the ``\textit{temp-attn.}" setting, we only finetune the SAWT-Attn. layers and freeze all other components. In the ``\textit{cross+temp-attn.}" setting, we jointly update the parameters of spatial-cross attn. and SAWT-Attn. layers. We observe that our default setting achieves the highest quality of video prediction among all these settings. Based on our default setting, further fine-tuning ``\textit{cross+temp-attn.}" causes the performance of Seer to drop a lot. These results suggest that the optimization of the FSText decomposer is strongly guided by the frozen conditional diffusion prior.

\subsection{Visual Analysis of Instruction Embedding}\label{sec:results:subins}
\begin{wrapfigure}[10]{r}{0.35\textwidth}
\centering
\vspace{-40pt}
\includegraphics[width=0.8\linewidth]{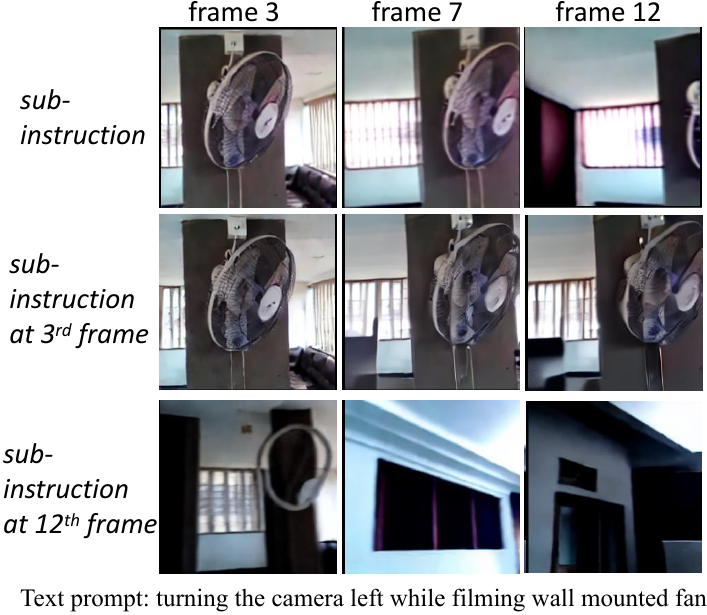}
\vspace{-10pt}
\caption{TVP results conditioned on frame-wise sub-instruction and constant clone of sub-instruction.}
\label{fig:sub_vis}
\end{wrapfigure}
To assess the impact of sub-instruction guidance on frame generation and unveil the implicit semantic information contained within a single sub-instruction at specific time steps, we compare text-conditioned video prediction (TVP) results conditioned on default frame-wise sub-instructions with those conditioned on the constant clone of a sub-instruction from the third or twelfth frame along the video axis, as depicted in Figure~\ref{fig:sub_vis}. Unlike the default frame-wise sub-instruction guidance, the sub-instruction clone from the twelfth frame tends to produce a transition from the reference frame to the video's termination state without temporal coherence. In contrast, the sub-instruction clone from the third frame tends to guide the motion until an intermediate state of the video, indicating that temporal sub-instructions provide proximate semantic guidance for motion at each time step. These findings underscore that sub-instructions align with the temporal sequence of the global instruction, offering fine-grained guidance across multiple steps. See more results in the Appendix~\ref{appendix:sec:crossatten}.


\subsection{Zero-shot Evaluation}\label{sec:ablate:gen}
\begin{wraptable}[3]{r}{0.38\textwidth}
\vspace{-45pt}
\centering\small
\setlength{\tabcolsep}{4pt}
\caption[gen]{\textbf{Zero-shot text-conditioned video prediction on EGO4D}}
\begin{tabular}{c|cc}
temp. attn. & FVD$\downarrow$ & KVD$\downarrow$\\
 \hline
 \textit{VideoFusion} & 618.6 & 1.85\\
 \textit{Seer(Ours)} &  301.7 & 0.55\\
\end{tabular}
\label{table:ablation:general}
\end{wraptable}
To assess the model's generalizability on an unseen dataset, we conducted a comparative analysis between our proposed Seer model and the current state-of-the-art baseline VideoFusion. Both models were fine-tuned on the Something Something-V2 dataset and evaluated on the EGO4D dataset, with the results summarized in Table~\ref{table:ablation:general}. Notably, Seer method outperformed VideoFusion, as evidenced by its superior performance in terms of FVD and KVD metrics. Drawing insights from these results, we can conclude that our default setup demonstrates strong generalizability while effectively adapting to the EGO4D dataset with higher-quality video generation.

\section{Conclusion}
In this paper, we propose Seer, a sample and computation-efficient model, for the challenging text-conditioned video prediction (TVP) task. We design a data and computation-efficient video network with Frame Sequential Text (FSText) Decomposer to inflate the pretrained text-to-image (T2I) stable diffusion models along the temporal axis. With the rich prior knowledge contained in pretrained T2I models and the well-designed architecture, Seer successfully generates high-quality videos by only fine-tuning the SAWT-Attn and FSText Decomposer, which significantly reduces the data and computation costs. The experiments illustrate our superior performance over all the recent models.

\section{Reproducibility Statement}
The main implementations of our proposed method are in Section~\ref{sec:impl} and ~\ref{sec:dataset}. In addition, the settings of the
experiments and hyper-parameters we choose are in Appendix~\ref{appendix:sec:impl}. And the implementation details are in Appendix~\ref{sec:appendix:impl_base}.

\section*{Acknowledgement}
This work is supported by the National Key R\&D Program of China (2022ZD0161700).  This work is also supported by the Ministry of Science and Technology of the People's Republic of China, the 2030 Innovation Megaprojects ``Program on New Generation Artificial Intelligence" (Grant No. 2021AAA0150000).

\bibliography{iclr2024_conference}
\bibliographystyle{iclr2024_conference}
\newpage
\appendix
\section{Appendix}
\section{Additional Experimental Results}

\subsection{Implementation Details of Baselines}\label{sec:appendix:impl_base}
We compare three baselines in our paper. For Tune-A-Video, to ensure a fair comparison, we use the pre-trained weight of Stable Diffusion-v1.5\footnote{https://github.com/CompVis/stable-diffusion} (same as our model) to initialize the UNet and we fine-tune the model with an image resolution of $256 \times 256$ on the training sets of Something Something-V2 (SSv2), Bridgedata and Epic-Kitchens-100 for 200k training steps. Similarly, we further fine-tune pre-trained VideoFusion on Something Something-V2, Epic-Kitchens-100, and BridgeData for 200k training steps. For MCVD, SimVP and PVDM, we train the model with an image resolution of $256 \times 256$, $64\times 64$, $256\times 256$ respectively on the training sets of SSv2, Epic-Kitchens-100, and Bridgedata for 300k training steps. For TATS and MAGE, we fine-tune the pre-trained UCF-101 model with an image resolution of $128 \times 128$ on the training sets of SSv2, Epic-Kitchens-100, and Bridgedata for 300k training steps.

\subsection{Evaluation Details and Results of UCF-101}\label{appendix:sec:ucf101}
Most prior text-conditioned video generation methods~\citep{hong2023cogvideo,vdm,makeavideo,magicvideo} evaluate their performance on the UCF-101~\citep{ucf101} benchmark. However, since our proposed method, Seer, is designed for text-conditioned video prediction (TVP) on task-level video datasets, the UCF-101 benchmark, which evaluates class-conditioned video prediction on random short-horizon video clips, is not an ideal evaluation benchmark for TVP. Nonetheless, in order to fairly compare these baselines, we still evaluate the class-conditioned video prediction performance of Seer on UCF-101. 

\paragraph{Settings}Specifically, we fine-tune our model with a video resolution of $16\times256\times256$ on UCF-101. Following the evaluation protocols of ~\citep{hong2023cogvideo}, Seer predicts the videos conditioned on 5 reference frames during fine-tuning and inference stage. We report FVD and Inception score (IS) metrics on the UCF-101 dataset~\citep{ucf101}. The IS is calculated by a C3D model\citep{c3d} that is pre-trained on the Sports-1M dataset~\citep{sports} and fine-tuned on UCF101. We follow the evaluation code of TGAN-v2~\citep{tganv2} to calculate IS metric. Following ~\citep{hong2023cogvideo,vdm,makeavideo}, we evaluate the FVD metric with 2,048 samples and IS metric with 100k samples in the validation set of UCF-101.

\paragraph{Results} Table~\ref{table:tvp:ucf} presents the class-conditioned video prediction results on UCF-101, demonstrating that Seer outperforms CogVideo~\citep{hong2023cogvideo} and MagicVideo~\citep{magicvideo}, but falls short of Make-A-Video~\citep{makeavideo}. Make-A-Video employs unlabelled video pre-training on temporal layers and achieves the best performance among all other methods. While Make-A-Video shows superior performance on FVD and IS, Seer has the potential to further improve its generation performance by addressing the following two limitations. First, Seer has not been pre-trained on video data. Second, Seer obtains latent vectors via a pre-trained 2D VAE, which has not been fine-tuned on UCF-101 and limits the video generation quality of Seer (with 259.4 FVD and 68.16 IS reconstruction quality). However, as we focus on the text-conditioned video prediction task, addressing the above limitations on UCF-101 is out of the scope of this paper.

\begin{table*}
\centering\small
\setlength{\tabcolsep}{5pt}
\caption{\textbf{ Class-conditioned video prediction performance on UCF-101} we evaluate the Seer on the UCF-101 with 16-frames-long videos. Ex.data indicates that the model has been pre-trained or fine-tuned on extra datasets.
}
\label{table:tvp:ucf}
\begin{tabular}{cccc|cc}
 \hline 
 Method & Ex.data & Cond. & Resolution & FVD$\downarrow$ & IS$\uparrow$\\
 \hline
MoCoGAN-HD~\citep{mocogan} & No & Class.  & $256\times 256$ & 700\tiny{$\pm$24} & 33.95\tiny{$\pm$0.25}\\
VideoGPT~\citep{videogpt} & No & No  & $128\times 128$ & - & 24.69\tiny{$\pm$0.30}\\
RaMViD~\citep{ramvid} & No & No  & $128\times 128$ & - & 21.71\tiny{$\pm$0.21}\\
StyleGAN-V~\citep{stylegan} & No & No  & $128\times 128$ & - & 23.94\tiny{$\pm$0.73}\\
DIGAN~\citep{digan} & No & No  & & 577\tiny{$\pm$22} & 32.70\tiny{$\pm$0.35}\\
TGANv2~\citep{tganv2} & No & Class.  & $128\times 128$ & 1431.0 & 26.60\tiny{$\pm$0.47}\\
VDM~\citep{vdm} & No & No  & $64\times 64$ & - & 57.80\tiny{$\pm$1.3}\\
TATS-base~\citep{tats} & No & Class.  & $128\times 128$ & 278\tiny{$\pm$11} & 79.28\tiny{$\pm$0.38}\\
MCVD~\citep{mcvd} & No & No  & $64\times 64$ & 1143.0 & -\\
LVDM~\citep{lvdm} & No & No  & $256\times 256$ & 372\tiny{$\pm$11} & 27\tiny{$\pm$1}\\
MAGVIT-B~\citep{magvit} & No & Class.  & $128\times 128$ & 159\tiny{$\pm$2} & 83.55\tiny{$\pm$0.14}\\
 \hline
VideoFusion~\citep{videofusion} & txt-video & Class.  & $128\times 128$ & 173 & 80.03\\
CogVideo~\citep{hong2023cogvideo} & txt-img \& txt-video & Class.  & $160\times 160$ & 626 & 50.46\\
Make-A-Video~\citep{makeavideo} & txt-img \& video & Class.  & $256\times 256$ & 81.25 & 82.55\\
MagicVideo~\citep{magicvideo} & txt-img \& txt-video & Class.  & & 699 & -\\
\textbf{Seer(Ours)} & txt-img & Class. & $256\times 256$ & 260.7 & 57.74\\
\hline 
\textbf{pre-trained VAE*} & - & - & $256\times 256$ & 259.4 & 68.16\\
 \hline
\end{tabular}
*we evaluate the reconstruction quality of pre-trained 2D VAE in this table, the pre-trained 2D VAE is initialized with the pre-trained weight from Stable Diffusion-v1.5 without extra fine-tuning.
\end{table*}

\subsection{Evaluation Results of Sampling Steps}
\begin{figure}
\centering
\includegraphics[width=1.0\linewidth]{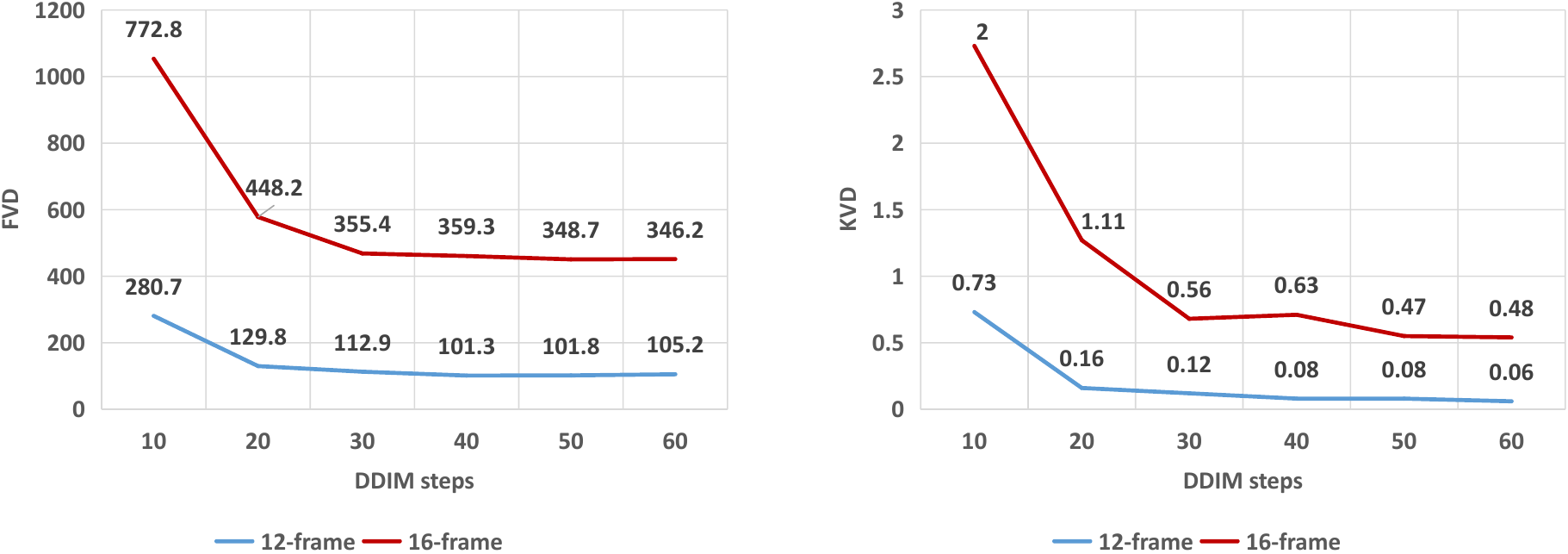}
\caption{Evaluation results of sampling 12-frame video and 16-frame video using 12-frame Seer model with DDIM sampling steps ranging from 10 to 60 on the Something-Something V2 dataset.}
\vspace{-8pt}
\label{fig:ddimvideostep}
\end{figure}

\textbf{Sampling steps:} To assess the influence of sampling steps on the quality of video predictions across varying sequence lengths, we conducted an evaluation on both 12-frame and 16-frame video predictions using a series of DDIM sampling steps (10, 20, 30, 40, 50, 60 DDIM steps). All generated outputs were sampled utilizing a 12-frame SSv2 fine-tuned model. The comparative results are presented in Figure~\ref{fig:ddimvideostep}. Notably, the 16-frame curve exhibits a more rapid decline from DDIM steps 10 to 20 compared to the 12-frame curve. As both curves progress beyond DDIM step 30, they tend to stabilize, showing marginal gains. These findings collectively underscore that increasing DDIM sampling steps notably enhances video quality for longer sequences (10 to 30 DDIM steps). However, the quality improvements of longer videos diminish as DDIM steps exceed 30.

\textbf{Video length:} Furthermore, we present the qualitative outcomes of the 16-frame video in Figure~\ref{fig:vis_16frame}. A comparison with the results of the 12-frame video in Figure~\ref{fig:ssv2mani} reveals that both the 16-frame and 12-frame videos can effectively capture task-described motion, as demonstrated in the example of "moving pen." However, the 16-frame video exhibits a gradual loss of appearance information from the reference frame and a decline in temporal consistency along the temporal axis with the increasing sequence length. Given that 16-frame videos are beyond the anticipated sequence length for a 12-frame model, enhancing the generation quality of the 16-frame video could be achieved through the training of a dedicated 16-frame Seer model.

\begin{figure}
\centering
\includegraphics[width=1.0\linewidth]{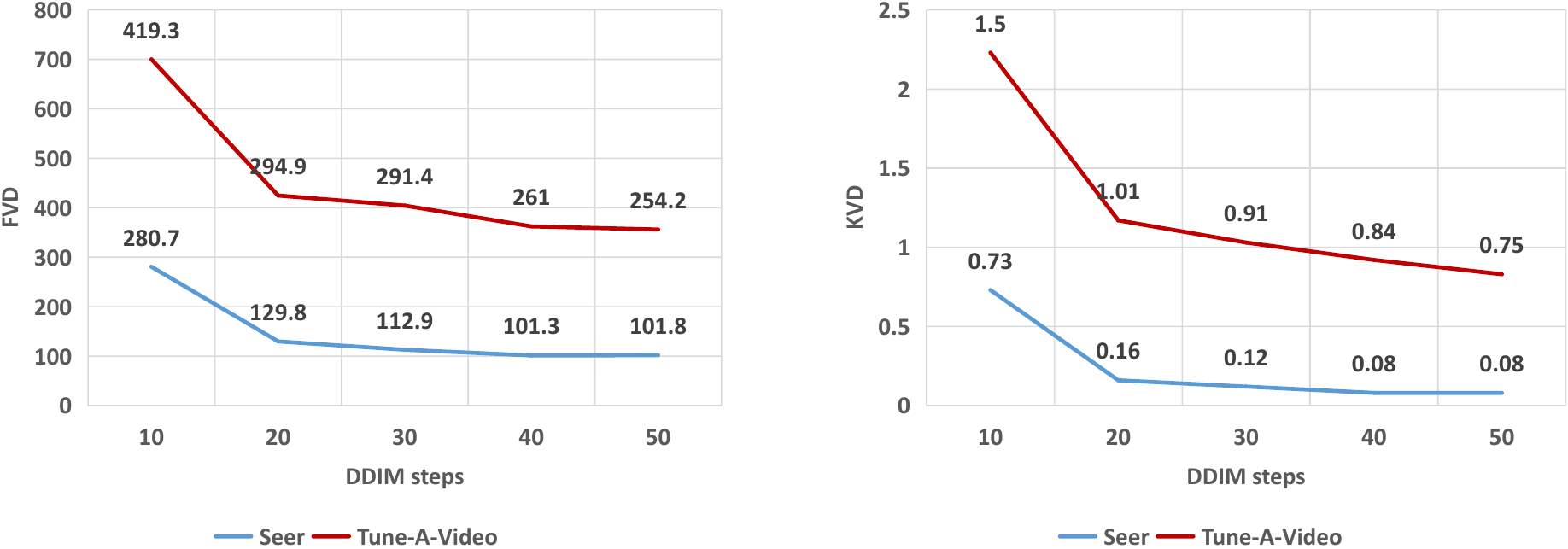}
\caption{Evaluation results of Seer and Tune-A-Video with DDIM sampling steps ranging from 10 to 50 on the Something-Something V2 dataset.}
\vspace{-8pt}
\label{fig:ddimstep}
\end{figure}

\begin{figure}[h]
\centering
\includegraphics[width=1.0\linewidth]{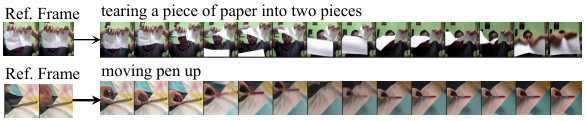}
\vspace{-18pt}
\caption{Visualization of 16-frame text-conditioned video prediction sampled by 12-frame Seer on SSv2.}
\label{fig:vis_16frame}
\end{figure}

\textbf{Methods comparison:} In addition, we explore the impact of fast sampling on generation results during evaluation by comparing Seer with Tune-A-Video.We apply a series of DDIM sampling steps (10, 20, 30, 40, 50 DDIM steps), as shown in Figure~\ref{fig:ddimstep}. Seer consistently outperformed Tune-A-Video in terms of both FVD and KVD, with improvements observed from 20 DDIM steps to 50 DDIM steps. Particularly noteworthy is Seer's advantage in video quality (280.7 FVD and 0.73 KVD) compared to Tune-A-Video (419.3 FVD and 1.5 KVD) when using only 10 DDIM steps, demonstrating Seer's ability to sample high-fidelity videos efficiently with minimal denoising steps.

\begin{wraptable}[8]{r}{0.35\textwidth}
\centering\small
\setlength{\tabcolsep}{4pt}
\vspace{-20pt}
\caption[temp]{\textbf{Training time of the variants of temporal attention on a 16-frame video}}
\begin{tabular}{c|c}
temp. attn. & (sec./iter.)\\
 \hline
 \textit{bi-direct.} & 2.35\\
 \textit{directed.} & 2.35\\
 \textit{autoreg.} & 5.50\\
 \textit{win-auto.}(Ours) & 2.40\\
\end{tabular}
\vspace{-10pt}
\label{table:speed_temp}
\end{wraptable}
\subsection{Computation Efficiency of Seer} 
To enhance the computational efficiency of Seer, both the inclusion of frozen 2D layers and the thoughtful design of temporal layers contribute significantly. A comprehensive evaluation of the computational cost for different temporal layer types, conducted on a single 24GB NVIDIA 3090 GPU, is presented in Table~\ref{table:speed_temp}. Remarkably, in comparison to plain autoregressive spatial-temporal attention (autoreg.), both bi-directional/directed temporal attention (bi-direct./directed.) and SAWT-Atten (win-auto.) substantially reduce computational overhead. Furthermore, SAWT-Atten demonstrates superior generation quality compared to bi-directional and directed temporal attentions, as evidenced by the ablation results in Section~\ref{sec:ablate:temp} of this paper.
\begin{table}[h]
\begin{minipage}{0.5\textwidth}
\centering\small
\vspace{-8pt}
\setlength{\tabcolsep}{3pt}
\caption[temp]{\textbf{Training time (time.) and GPU memory (Mem.) consumption of the models (16-frame)}}
\vspace{-5pt}
\begin{tabular}{c|c|c|c}
\multirow{2}{*}{model} & \multirow{2}{*}{2D. frozen} & time. & Mem.\\
 &  & (sec./iter.) &(GB)\\
 \hline
 \textbf{\textit{Seer (Ours)}} & Yes & 0.75 & 24.7\\
 \textit{Seer }& No & 0.96 & 39.2\\
 \textit{VideoFusion.}& No & 1.07 & 45.0\\
\end{tabular}
\vspace{-8pt}
\label{table:speed_seer}
\end{minipage}
\begin{minipage}{0.5\textwidth}
\vspace{-10pt}
\centering\small
\setlength{\tabcolsep}{3pt}
\caption[temp]{\textbf{Training time (time.) and GPU memory (Mem.) consumption of the models (16-frame)( $\geq$ 90\% GPU memory usage)}}
\begin{tabular}{c|c|c|c}
\multirow{2}{*}{model} & \multirow{2}{*}{2D. frozen} & time. & Mem.\\
 &  & (sec./iter.) &(GB)\\
 \hline
 \textbf{\textit{Seer (Ours)}} & Yes & 3.10 & 72.9\\
 \textit{Seer }& No & 6.89 & 75.8\\
 \textit{VideoFusion.}& No & 7.63 & 78.7\\
\end{tabular}
\vspace{-8pt}
\label{table:speed_seer_GPU}
\end{minipage}
\end{table}
In addition, a comparison involving our ablated setting and the baseline method VideoFusion~\citep{videofusion}, which shares a network design based on the Stable Diffusion U-Net, is presented. In our proposed Seer setting, the 2D spatial layers of the U-Net are frozen during fine-tuning, while the ablated setting maintains all 2D spatial layers trainable. GPU memory consumption and training time for different models were assessed running with a single 80GB NVIDIA A800 GPU. These results are available in Table~\ref{table:speed_seer}, and highlight the proposed Seer setting with frozen 2D layers exhibiting approximately half the GPU memory consumption (24.7GB) compared to Seer without frozen 2D layers (39.2GB) and VideoFusion (45.0GB). To ensure a fair comparison of training speed among various settings on the single A800 GPU, we conducted additional assessments of models with GPU memory usage exceeding 90\%, as detailed in Table~\ref{table:speed_seer_GPU}. The results in the table reveal a notable advantage in computation efficiency for the proposed Seer setting with frozen 2D layers, exhibiting reduced training time (3.10 sec/iter) compared to Seer without frozen 2D layers (6.89 sec/iter) and VideoFusion (7.63 sec/iter). These findings firmly establish the enhanced computational efficiency of Seer relative to baseline models.

\subsection{Additional Ablation Results}\label{appendix:sec:fstext}
\paragraph{FSText layer depth}In this section, we additionally investigate the impact of FSText Decomposer's layer depth in Table~\ref{table:ablation:layer}. Our default setting (8-layer FSText Decomposer) outperforms shallower models (2-layer and 4-layer) in terms of FVD. Though the 4-layer model shows a marginal advantage over the 8-layer model in terms of KVD, our experiments indicate that the 8-layer FSText Decomposer shows a remarkable advantage on FVD metrics and exhibits robustness in text-video alignment. Therefore, we adopt the 8-layer FSText Decomposer as the default setting for Seer.
\paragraph{FSText componenets}To evaluate the impact of individual attention layers within the FSText decomposer, we conducted an ablation study on the FSText component, as presented in Table 8. In this study, we ablate the temporal attention layer, labeled as "Temp.," and the text-sequential attention layer, denoted as "Seq.," within the FSText network. To maintain consistency in model size across different settings, we replaced the ablated component with a Cross-Attention layer. The results, shown in Table~\ref{table:fstext_component}, highlight the superiority of our proposed setting, which integrates both text-sequential-attention layers and temporal attention layers. Our proposed setting outperforms the other two settings, underscoring the significant attributes of Text-Sequential-Attention layers and Temporal Attention layers to capture text contextual information and model temporal dependencies, collectively enhancing the overall performance of the FSText decomposer.
\paragraph{Additional ablation on BridgeData}To validate the robustness and consistency of Seer ablation results across different datasets, we conducted additional experiments on the BridgeData dataset, extending our analysis from the Something Something-V2 (SSv2) dataset. The corresponding ablation studies on Seer fine-tuning settings are presented in Table~\ref{table:ablate_finetune_bridge}, while the temporal layer settings are detailed in Table~\ref{table:ablate_tempattn_bridge}. These results mirror the ablation outcomes reported in Table~\ref{table:ablation:finetune} and Table~\ref{table:ablation:tempattn} for the SSv2 dataset. Notably, the consistent improvement observed in both Seer fine-tuning and temporal layer ablation across different datasets, as demonstrated in Table~\ref{table:ablate_finetune_bridge} and Table~\ref{table:ablate_tempattn_bridge} on the BridgeData dataset, demonstrates the robustness of the Seer component design.
\begin{table}
\begin{minipage}{0.5\textwidth}
\centering\small
\vspace{-20pt}
\setlength{\tabcolsep}{5pt}
\caption{\textbf{Layer depth in FSText. (SSv2)}.}
\label{table:ablation:layer}
\begin{tabular}{c|cc}
num. layers.& FVD$\downarrow$ & KVD$\downarrow$\\
 \hline
 2 & 238.6 & 0.51\\
 4 & 229.7 & 0.23\\
8(Ours) &112.9 & 0.12\\
\end{tabular}
\end{minipage}
\begin{minipage}{0.5\textwidth}
\vspace{-30pt}
\centering\small
\setlength{\tabcolsep}{4pt}
\caption{\textbf{Components in FSText Decomposer (SSv2)}. Our settings are marked in \colorbox{baselinecolor}{gray}}
\label{table:fstext_component}
\begin{tabular}{cc|cc}
Temp. & Seq. & FVD$\downarrow$ & KVD$\downarrow$\\
 \hline
 \checkmark & & 125.8 & 0.13\\
 & \checkmark & 127.7 & 0.14\\
\baseline{\checkmark} & \baseline{\checkmark} & \baseline{112.9} & \baseline{0.12}\\
\end{tabular}
\vspace{-8pt}
\end{minipage}
\end{table}

\begin{table}
\begin{minipage}{0.5\textwidth}
\centering\small
\vspace{-10pt}
\setlength{\tabcolsep}{5pt}
\caption[temp]{\textbf{Ablation study of temporal attention (BridgeData)}}
\begin{tabular}{c|cc}
temp. attn. & FVD$\downarrow$ & KVD$\downarrow$\\
 \hline
 \textit{bi-direct.} & 284.5 & 0.71\\
 \textit{directed.} & 258.0 & 0.64\\
 \textit{autoreg.} & 261.5 & 0.83\\
 \textit{win-auto.}(Ours) &246.3 & 0.55\\
\end{tabular}
\label{table:ablate_tempattn_bridge}
\end{minipage}
\begin{minipage}{0.5\textwidth}
\vspace{-20pt}
\centering\small
\setlength{\tabcolsep}{4pt}
\caption[fine]{\textbf{Ablation study of Fine-tune settings (BridgeData)}}
\begin{tabular}{cc|cc} 
fine-tune& FSText. & FVD$\downarrow$ & KVD$\downarrow$\\
 \hline
 \textit{temp-attn.} & & 410.7 & 0.97\\
 \textit{cross}+\textit{temp-attn.} & & 319.9 & 1.01\\
 \textit{temp-attn.}(Ours) & \checkmark &246.3 & 0.55\\
 \textit{cross}+\textit{temp-attn.} & \checkmark & 2058.4 & 9.43\\
\end{tabular}
\vspace{-8pt}
\label{table:ablate_finetune_bridge}
\end{minipage}
\end{table}

\begin{figure}[h]
\centering
\includegraphics[width=1.0\linewidth]{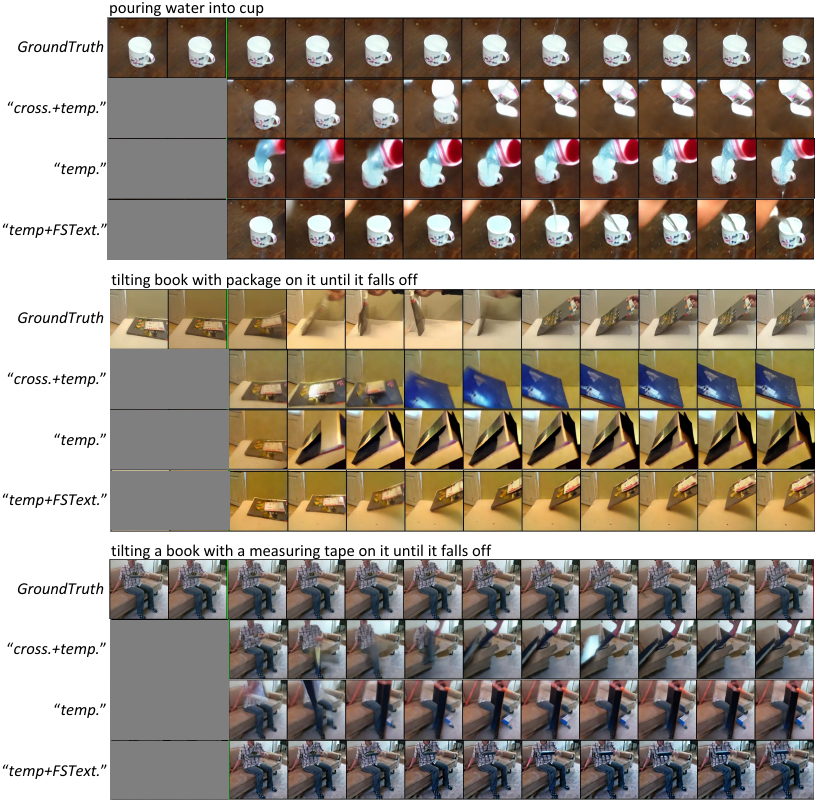}
\caption{Additional qualitative results of fine-tuning ablation. \textit{“temp+FSText.”} is our default setting.}
\vspace{-8pt}
\label{fig:ablate:finetune}
\end{figure}

\paragraph{Qualitative results of fine-tuning ablation} We conduct a qualitative analysis of various fine-tune settings. We provide additional visualizations of Fine-tune Setting ablation in Section 5.5 of the main paper. Figure~\ref{fig:ablate:finetune} shows the results of different settings. Among these settings, our default setting \textit{``temp+FSText"} stands out as it preserves a higher-level temporal consistency in video prediction starting from reference frames and also delivers superior text-based video motion compared to the other fine-tune settings. 

\subsection{Evaluation of Policy Learning on Seer}\label{appendix:sec:policylr}
To investigate whether Seer can help policy learning, we choose the UniPi~\citep{dai2023unipi} as our baseline in the simulated robotics environment Meta-World~\citep{metaworld}, which generates videos from the initial state and infers actions from the adjacent frames via a pre-trained inverse-dynamics model. Specifically, we distill a policy model from the videos generated by the Seer and the labeled actions from the pretrained inverse-dynamics model. We choose 3 tasks and use 10 in-domain videos for each task to fine-tune the Seer. And we make comparisons between
\textbf{a) Policy a:} Distilled policy from the dataset generated by the Fine-tuned Seer model (1000 generated videos for each task).
\textbf{b) Policy b:} Distilled policy from the 10 in-domain videos for each task. The results are shown in Table~\ref{table:policy}, where we find that compared to utilizing the given 10 in-domain videos to generate policy, fine-tuning the Seer with them and generating more videos can be better because it can acquire more scalable data, which is of comparable quality. Therefore, the videos generated by Seer can help policy learning in simulated robotics tasks in a way. The visualization of Seer within a robot simulation environment is presented in Section~\ref{appendix:sec:robotvision}
\begin{wraptable}[7]{r}{0.45\textwidth}
\centering\small
\setlength{\tabcolsep}{4pt}
\vspace{-8pt}
\caption[policy]{\textbf{Success rate of distilled policy}}
\begin{tabular}{c|c|c}
tasks & \textbf{policy a}& policy b\\
 \hline
 \textit{button-press-topdown-v2} & \textbf{0.45} & 0.4\\
\textit{drawer-close-v2} & \textbf{0.1} & 0.0\\
\textit{drawer-open-v2} & \textbf{0.05} & 0.0\\
\end{tabular}
\label{table:policy}
\end{wraptable}
\clearpage
\section{Implementation Details}\label{appendix:sec:impl}

\subsection{Fine-tuning and Sampling}\label{sec:finetuneparam}
 In this section, we list the hyperparameters, fine-tuning details, sampling details, and hardware information of our model in Table~\ref{table:hyperparam:finetune}.
 
\subsection{Architecture information}\label{sec:arch}
In this section, we list the hyperparameters of 3D U-Net in Table~\ref{table:3dunet} and hyperparameters of FSText Decomposer in Table~\ref{table:hyperparam:fstext}.

\begin{table}[h]
\centering\small
\caption{Hyperparameters and details of Fine Tuning/Inference}
\label{table:hyperparam:finetune}
\begin{tabular}{c|cc}
\hline
param. & value\\
\hline
optim. & AdamW\\
Adam-$\beta_1$ &  0.9\\
Adam-$\beta_2$ &  0.99\\
Adam-$\epsilon$ &  $1e^{-8}$\\
weight decay &  $1e^{-2}$\\
lr &  $1.024e^{-4}$\\
end lr & 0.0\\
lr sche. & cosine\\
noise sche. & cosine\\
train batch size& 1/GPU\\
grad. acc.& 2\\
warmup steps& 10k\\
resolution& $256 \times 256$\\
train. steps & 200k\\
train. hardware & 4 RTX 3090\\
val. batch size& 2/GPU\\
sampler& DDIM\\
sampling steps & 30\\
guidance scale & 7.5\\
\hline
\end{tabular}
\end{table}

\begin{table}[h]
\centering\small
\caption{Hyperparameters of 3D U-Net}
\label{table:3dunet}
\begin{tabular}{c|cc}
\hline
hyperparam. & value\\
\hline
input/output channels &  4\\
Base channels & 320\\
Channel multipliers&  1,2,4,4\\
3D Downsample blocks &  4\\
3D Upsample blocks &  4\\
Number of layers (per block) &  2\\
\hline
Modules of layer & 3D ResnetBlock\\
 & Spatial-cross Atten.\\
 & SAWT-Atten.\\
 & Down./Up. 3D ResnetBlock\\
\hline
Dimension of atten. heads &  8\\
activation function &  SiLU\\
Dimension of cross-atten. &  768\\
\hline
\end{tabular}
\end{table}

\begin{table}[h]
\centering\small
\caption{Hyperparameters of FSText Decomposer}
\label{table:hyperparam:fstext}
\begin{tabular}{c|cc}
\hline
hyperparam. & value\\
\hline
learnable tokens channels &  768\\
output channels &  768\\
Base channels & 768\\
Number of layers &  8\\
\hline
Modules of layer & Seq-cross Atten.\\
 & Feedforward\\
 & Directed temporal Atten.\\
 & Feedforward\\
\hline
Number of atten. heads &  8\\
Dimension of cross-atten. &  768\\
\hline
\end{tabular}
\end{table}
\newpage
\section{Visualization} 

\subsection{Additional qualitative results} 
We provide additional visualization on Something-Something v2 (SSv2) of our text-conditioned video prediction in Figure~\ref{fig:ssv2pred}, and text-conditioned video prediction/manipulation results in Figure~\ref{fig:ssv2mani}. Additionally, we provide the visualization on BridgeData of text-conditioned video prediction in Figure~\ref{fig:bridgepred} and text-conditioned video prediction/manipulation in Figure~\ref{fig:bridgemani}. We also provide the visualization results of text-conditioned video prediction on Epic-Kitchens-100 in Figure~\ref{fig:epicpred}.
\begin{figure}[h]
\centering
\includegraphics[width=1.0\linewidth]{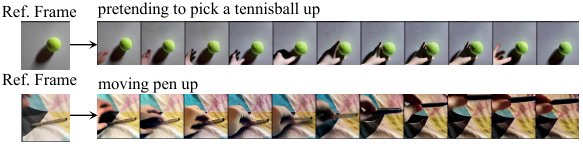}
\vspace{-10pt}
\caption{Additional visualization results of 12-frame Text-conditioned video prediction (reference frame=1) on SSv2 dataset.}
\label{fig:sub_instruct_appendix}
\end{figure}
\begin{wrapfigure}[14]{r}{0.5\textwidth}
\centering
\vspace{10pt}
\includegraphics[width=1.0\linewidth]{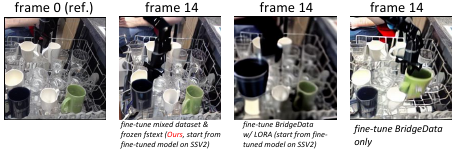}
\vspace{-20pt}
\caption{The visualization of comparison with different fine-tune settings on BridgeData (given text instructions "pick up black mug", where "black" text prompt is unseen in BridgeData). Compare to other two settings, our default setting successfully recognizes "black mug" and generates RGB frame with higher fidelity}
\label{fig:generlize}
\end{wrapfigure}
\subsection{Generalizability of Seer On Downstream Tasks}\label{sec:generlize}
To thoroughly assess Seer's adaptability on downstream tasks such as video manipulation on BridgeData, we conducted an investigation into diverse fine-tuning strategies on this dataset. Our default setting proceeded to fine-tune only the temporal block of 3D U-Net, freezing the FSText decomposer on a mixed dataset including both the pre-training video dataset and the task-specific downstream dataset. Additionally, we compare our default setting with LORA modules~\citep{lora}, which are integrated into each temporal layer of the UNet architecture and throughout the entire FSText decomposer.  As depicted in Figure~\ref{fig:generlize}, both our proposed setting and the LORA setting have effectively identified the previously unseen "black mug" in the training set of BridgeData. Besides, our proposed setting maintains a higher background consistency from the reference frame (frame 0) and generates future frames with superior fidelity when compared to the LORA setting. We also provide additional visualization in Appendix~\ref{appendix:sec:gen}.
\subsection{Additional Visualization of Sub-Instruction Embedding}\label{appendix:sec:crossatten} 
We present additional visualizations comparing frame-specific sub-instructions to those constant clones from the third or twelfth frame along the temporal axis (Figure~\ref{fig:sub_instruct_appendix}). These visualizations reveal that each frame's sub-instruction represents the motion at its corresponding time step. For example, in the case of "pushing iphone adapter from left to right" the default sub-instruction guides the entire instructed behavior with continuous temporal motion, while the sub-instruction embedding from the third frame directs the motion to a halfway point in the case of pushing the iphone adapter to right. In contrast, utilizing the sub-instruction embedding from the twelfth frame leads to frames that are disconnected from the reference frame's scene, resulting in the absence of transition between frames. This observation demonstrates that sub-instructions closely follow the temporal sequence of the global instruction, providing precise guidance across multiple steps.

\subsection{Additional Visualization of Video Manipulation With Unseen Objects And Zero-shot Video Manipulation}\label{appendix:sec:gen}
We also include visualizations of video manipulation with unseen objects in BridgeData and zero-shot generation on EGO4D. These experiments involved fine-tuning the SSv2 video model on a mixed dataset consisting of SSv2 and BridgeData. The fine-tuning process lasted for 800k iterations, employing a learning rate of $4.096e^{-5}$. Subsequently, the model's performance was evaluated on Bridgedata (as shown in Figure~\ref{fig:generlize_appendix}(a)) and the EGO4D dataset (illustrated on the right side of Figure~\ref{fig:generlize_appendix}(b)), where we sampled 15 frames with one reference frame. In the evaluation on BridgeData, Seer demonstrated the ability to recognize previously unseen objects such as "red plate" and "cabbage" and successfully performed a "pick up" motion, leveraging prior knowledge learned from SSv2. In certain scenarios, such as picking up the plate, where Seer has not encountered the specific action during training, the generated video depicted some limitations in the interaction between the robotic arm and the plate. Regarding the evaluation on EGO4D, although Seer had not been fine-tuned on this dataset, it exhibited the capability to accurately identify objects in the EGO4D environment, such as "laptop" and "book," and execute actions based on prior knowledge acquired from observing human activities. However, Seer still faced challenges in predicting future frames based on the understanding of the scene in the reference frame. For instance, in the case of "closing a book," Seer tended to generate a hand outside the camera view instead of manipulating the object with the main body, such as the hand holding the book, within the scene.

\clearpage
\begin{figure}
\centering
\includegraphics[width=0.9\linewidth]{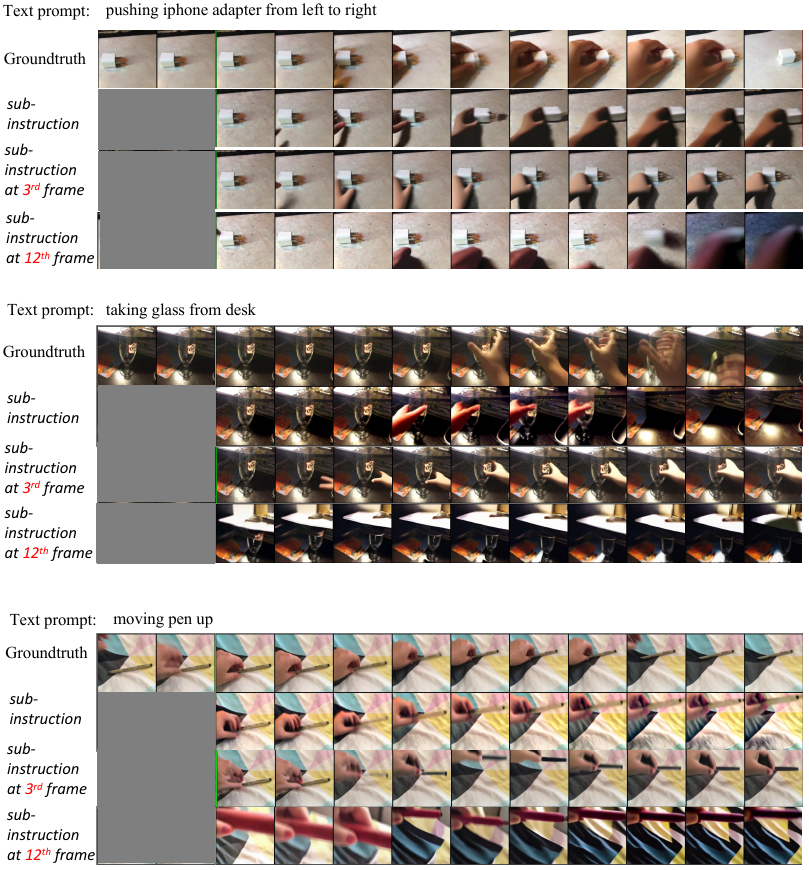}
\vspace{-10pt}
\caption{Additional visualization results of Seer's Text-conditioned video prediction conditioned on frame-wise sub-instruction and duplicate sub-instruction at the third/twelfth frame.}
\label{fig:sub_instruct_appendix}
\end{figure}

\begin{figure}
\centering
\includegraphics[width=1.0\linewidth]{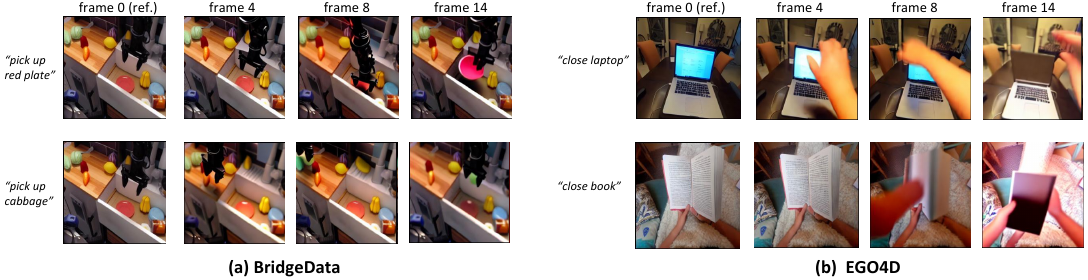}
\vspace{-20pt}
\caption{Additional visualization of Seer generalizability evaluation on unseen dataset EGO4D (b) and BridgeData (a), where "cabbage" and "red plate" are unseen objects in the training set of BridgeData.}
\label{fig:generlize_appendix}
\end{figure}
\clearpage
\subsection{Visualization of Robot Simulation Environment}\label{appendix:sec:robotvision}
\begin{figure}[h]
\centering
\includegraphics[width=0.85\linewidth]{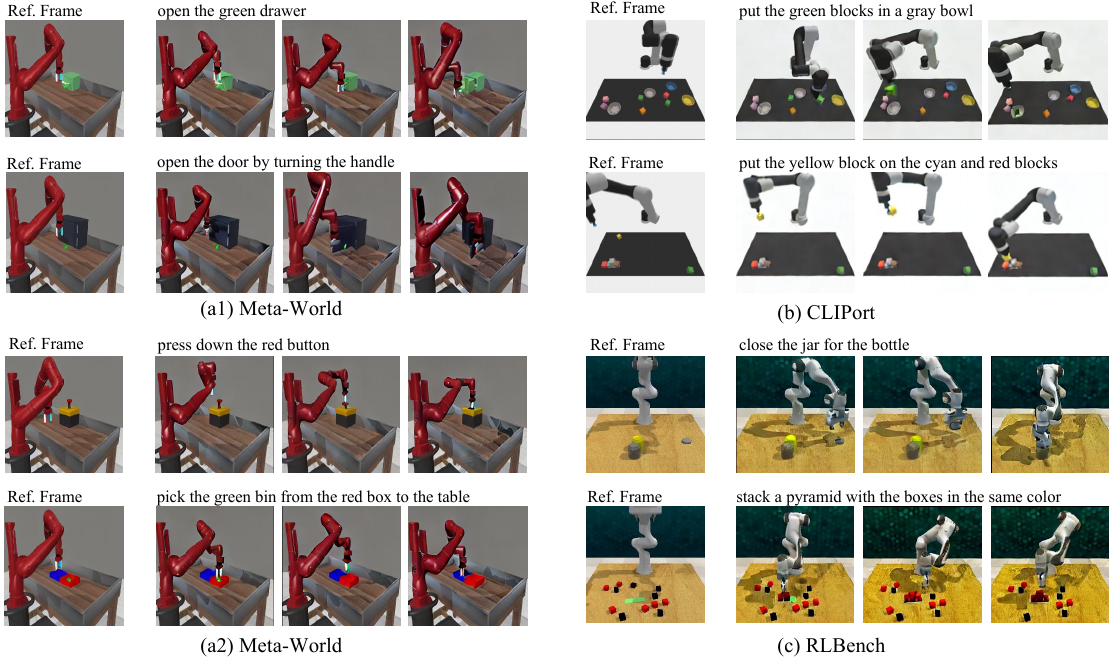}
\caption{Text-conditioned video prediction of Seer (conditioned on first frame) on Meta-World (a1-2), CLIPort (b), and RLBench (c) datasets}
\label{fig:robot_manipulate}
\end{figure}

To investigate the applicability of Seer in robot environments, we assess its performance on robot simulation datasets, including Meta-World~\citep{metaworld}, CLIPort~\citep{cliport}, and RLBench~\citep{rlbench}. Beginning with the SSv2-finetuned Seer model, we further fine-tune it on a mixed dataset comprising  Something
Something-V2 (SSv2), Bridgedata, 236 MetaWorld video clips, 785 CliPort video clips, and over 2000 video clips from RLBench for 80,000 steps, using a learning rate of 4e-5. Throughout the fine-tuning process, all 2D layers of the 3D inflated U-Net and the FSText decomposer remain frozen. Figure~\ref{fig:robot_manipulate} presents visualizations of Seer on these robot simulation datasets. In these visualizations, Seer successfully generates coherent videos with continuous motion trajectories aligned with language instructions. In the CLIPort video clips, Seer accurately detects small target objects in complex scenes and places them in the correct positions as instructed by the task descriptors. Moreover, in the RLBench videos, Seer demonstrates the ability to perform multi-step tasks such as "stacking a pyramid with the boxes." These observations highlight Seer's adaptability to a multi-task environment, encompassing human video, robot manipulation, and simulation scenarios, while maintaining robust temporal alignment with task descriptors. We conduct an assessment of Seer performance on policy learning, and a further analysis is presented in Section~\ref{appendix:sec:policylr}.

\subsection{Long Video Prediction}
To evaluate Seer's ability to generate extended video frames during the inference stage, we conducted a qualitative test on long video prediction using a 12-frame SSv2 fine-tuned Seer model. Our video sampling approach involves the sequential generation of video clips, where the first 12 frames are conditioned on 2 reference frames, and the last two frames of the first clip serve as the condition for the subsequent clip. The concatenation of these clips results in a long video. A crucial element for aligning long video frames with language is the presence of a prolonged sequence of sub-instructions. To achieve this, we employ two text conditioning strategies for expanding the sub-instruction embedding along the frame axis, enabling the iterative sampling of longer frames.

\begin{figure}[h]
\centering
\includegraphics[width=0.85\linewidth]{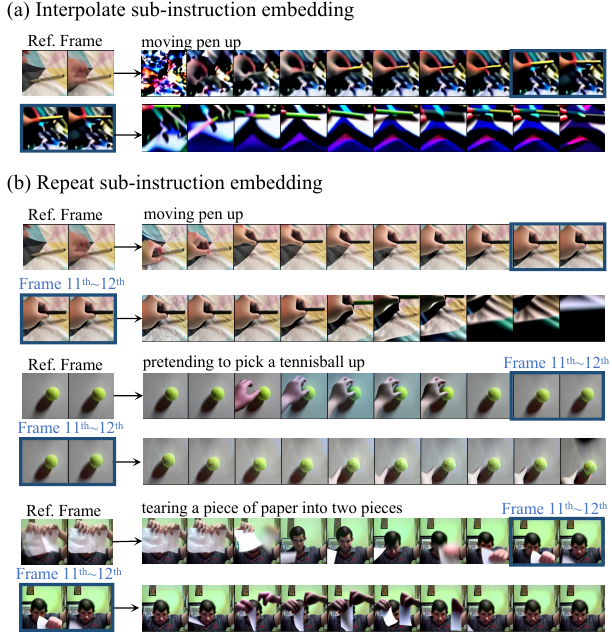}
\caption{22-frame video prediction conditioned on 2 reference frames using a 12-frame Seer model on SSv2. (a) Interpolation of 12-frame subinstruction embeddings to a 22-frame sequence. (b) Repetition of 12-frame subinstruction embeddings for the second video clip prediction from the eleventh frame to the twenty-second frame.}
\label{fig:longvideo}
\end{figure}

The first strategy involves interpolating sub-instruction embedding along the frame axis, while the second strategy entails repeating the sub-instruction embedding from the first video clip to guide the second clip. In Figure~\ref{fig:longvideo}, we compare these two strategies and observe that direct interpolation (Figure~\ref{fig:longvideo} (a)) tends to degrade overall generation quality, introducing unexpected noise. Conversely, utilizing the same sub-instruction (Figure~\ref{fig:longvideo} (b)) can maintain coherent motion, persisting until the target object is no longer present or the current motion is in a terminated state. Although this strategy facilitates the generation of coherent movements, it is distinct from a simple upsampling of video clips along the frame axis. It is noteworthy that all results were obtained using the 12-frame Seer model.

Intuitively, appending an additional upsampler network could potentially enhance Seer's ability to expand video frames, which causes extra computational costs. Observing the results of the evaluation, we believe that expanding the frame length of generated subinstruction embeddings during the training stage represents a promising direction for enabling Seer to generate longer video frames in multi-steps without incurring additional computational overhead.

\clearpage
\subsection{Failure Case}
\begin{figure}[h]
\centering
\includegraphics[width=1.0\linewidth]{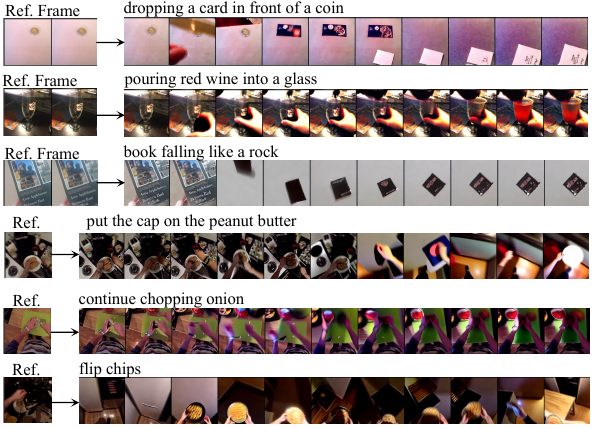}
\vspace{-18pt}
\caption{Seer failure cases on Something
Something-V2 (row 1,2,3) and Epic-Kitchens-100 (row 4,5,6)}
\label{fig:fail_case}
\end{figure}

In this section, we present instances where Seer encounters challenges in handling environmental motion in human-generated videos. The generated videos highlight situations such as "dropping a card in front of a coin" and "book falling like a rock" (refer to Figure~\ref{fig:fail_case}), where Seer successfully predicts task-descriptive motions like "dropping" and "falling" and correctly identifies text-described objects such as "card" and "book." However, the generated future frames fall short in capturing appearance consistencies, such as the color of the card in the previous frame and the cover of the book in the reference environment. In the scenario of "pouring red wine into a glass," Seer tends to generate a wine glass based on its knowledge of pouring red wine but overlooks the transition distribution from the reference environment.

Notably, in the Epic-Kitchens-100 dataset, where scene transitions are prevalent, Seer exhibits a preference for predicting camera pose movements and generating novel views of the environment, reflecting its imaginative capabilities. However, these outcomes extend beyond the scope of Seer's primary objective, which is to learn human behavior. Consequently, addressing challenges such as filtering out irrelevant background information, including camera pose and object occlusion, and refining Seer's awareness of temporal motion becomes imperative for its adaptation to learning from internet videos. 

\newpage
\section{Human Evaluation Details}\label{appendix:sec:humaneval} 
\begin{figure}[h]
\centering
\includegraphics[width=0.7\linewidth]{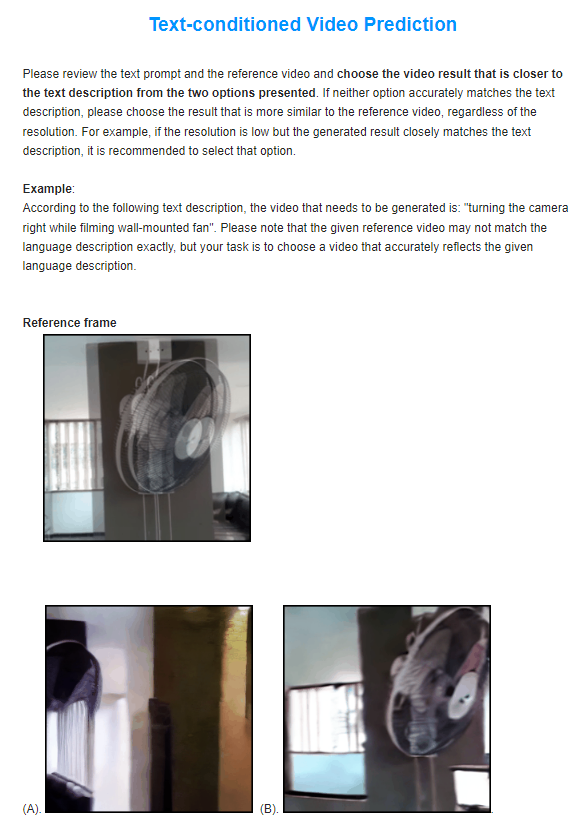}
\caption{Screenshot of a questionnaire example shown to human evaluators.}
\vspace{-18pt}
\label{fig:humanevalexp}
\end{figure}

\begin{figure}
\vspace{-38pt}
\centering
\includegraphics[width=0.65\linewidth]{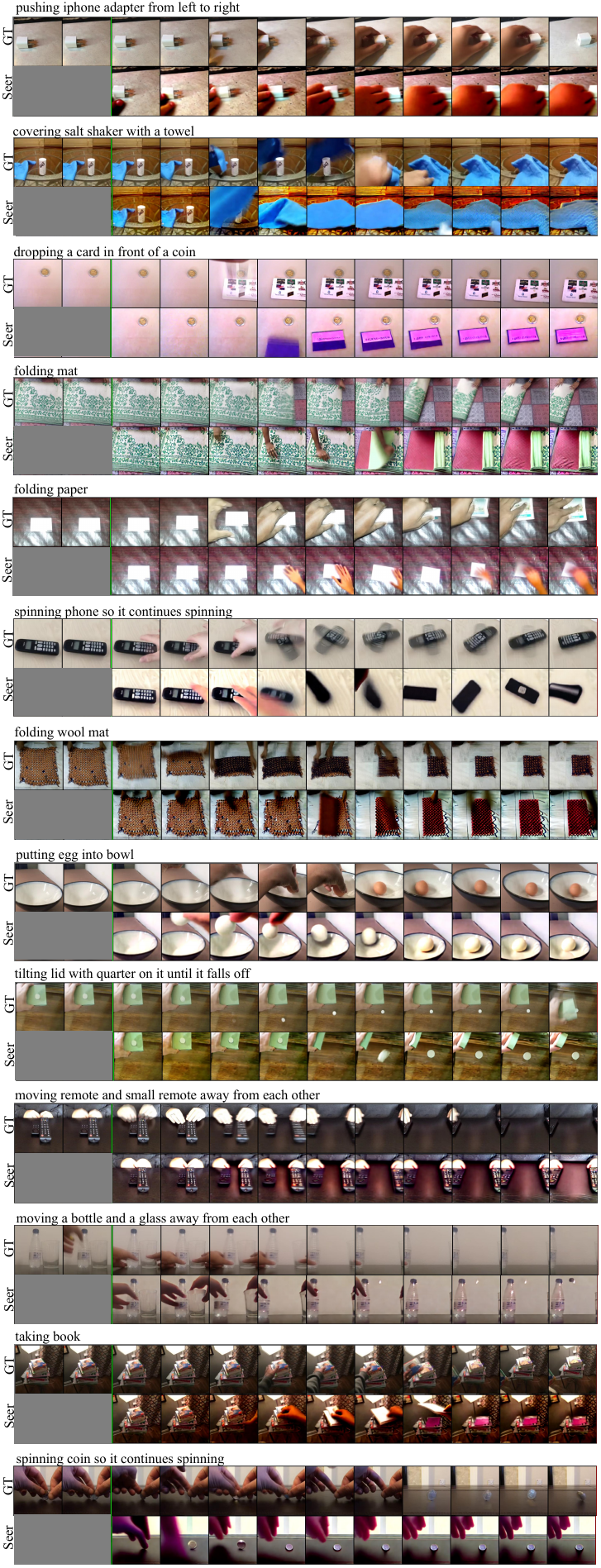}
\vspace{-8pt}
\caption{Text-conditioned video prediction of Seer on SSv2.}
\label{fig:ssv2pred}
\end{figure}

\begin{figure}
\vspace{-38pt}
\centering
\includegraphics[width=0.65\linewidth]{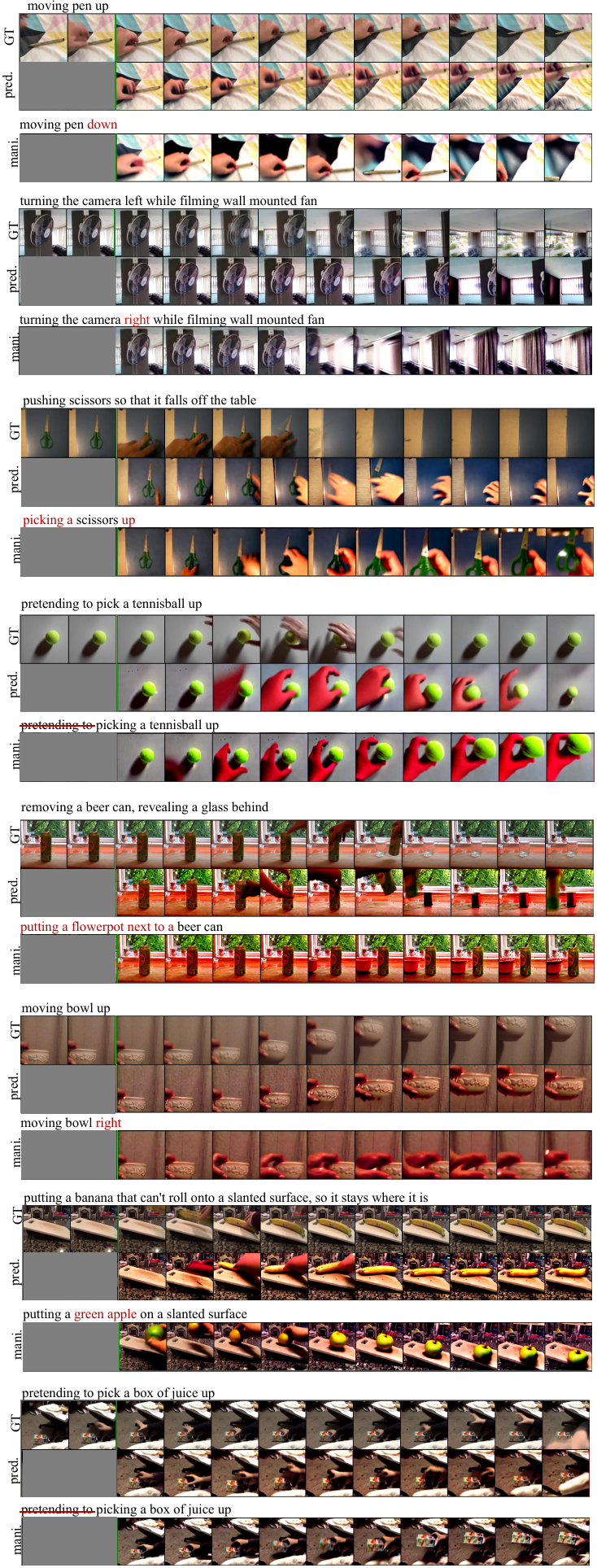}
\vspace{-8pt}
\caption{Text-conditioned video prediction/manipulation of Seer on SSv2, where ``pred." refers to prediction, ``mani." refers to manipulation.}
\label{fig:ssv2mani}
\end{figure}

\begin{figure*}
\centering
\includegraphics[width=0.9\linewidth]{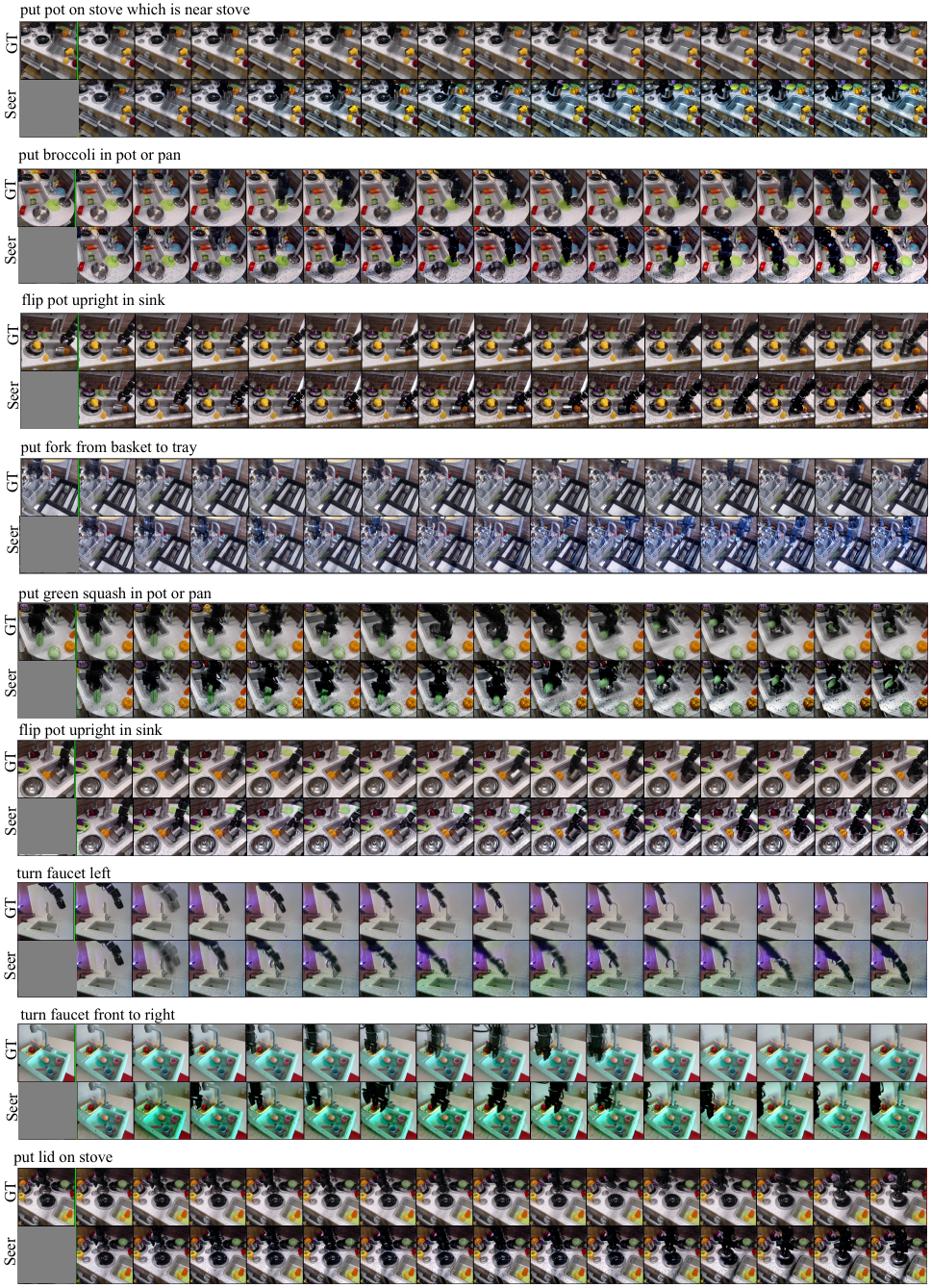}
\caption{Text-conditioned video prediction of Seer on BridgeData.}
\vspace{-8pt}
\label{fig:bridgepred}
\end{figure*}

\begin{figure*}
\centering
\includegraphics[width=1.0\linewidth]{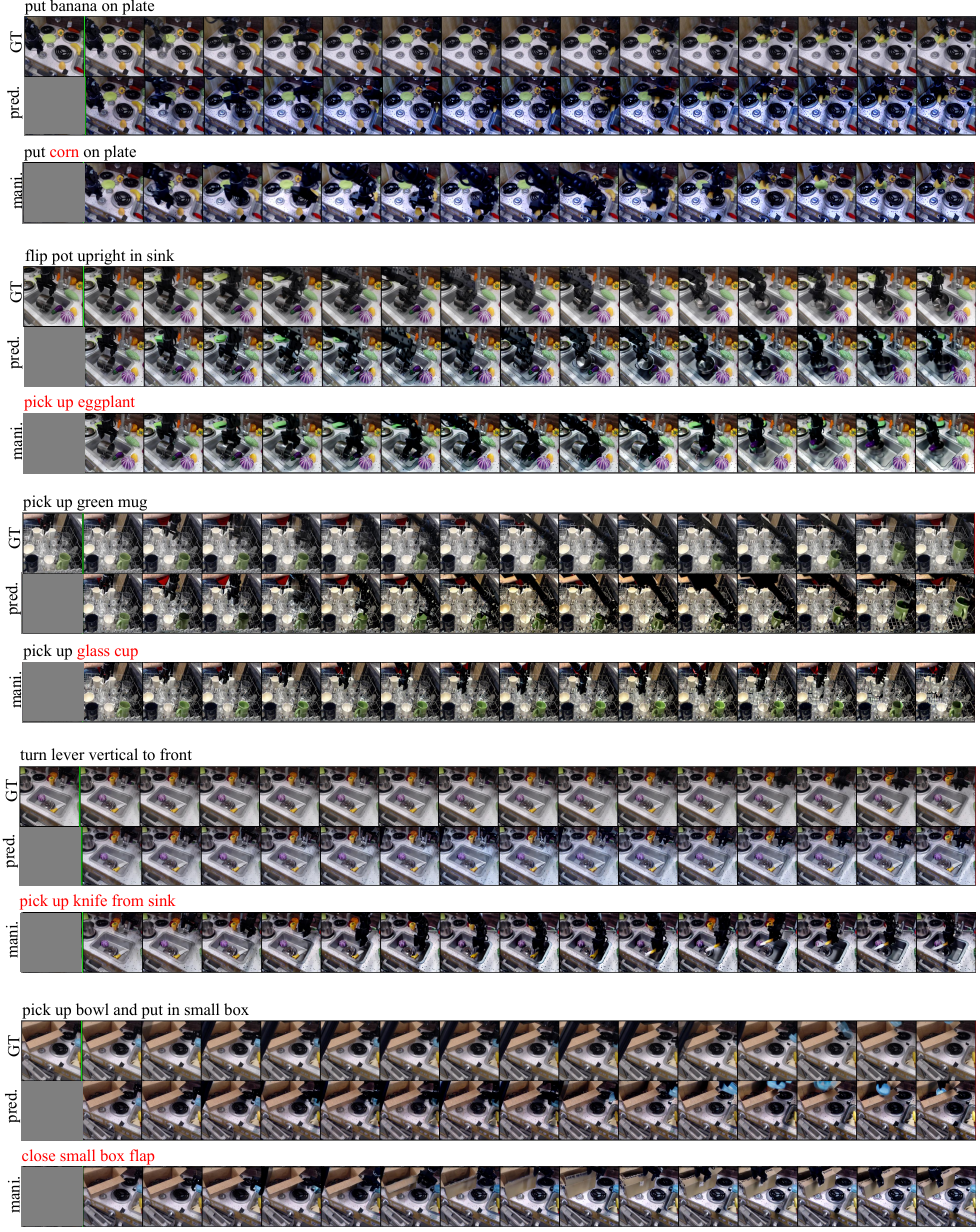}
\caption{Text-conditioned video prediction/manipulation of Seer on BridgeData, where ``pred." refers to prediction, ``mani." refers to manipulation.}
\vspace{-8pt}
\label{fig:bridgemani}
\end{figure*}

\begin{figure*}
\centering
\includegraphics[width=1.0\linewidth]{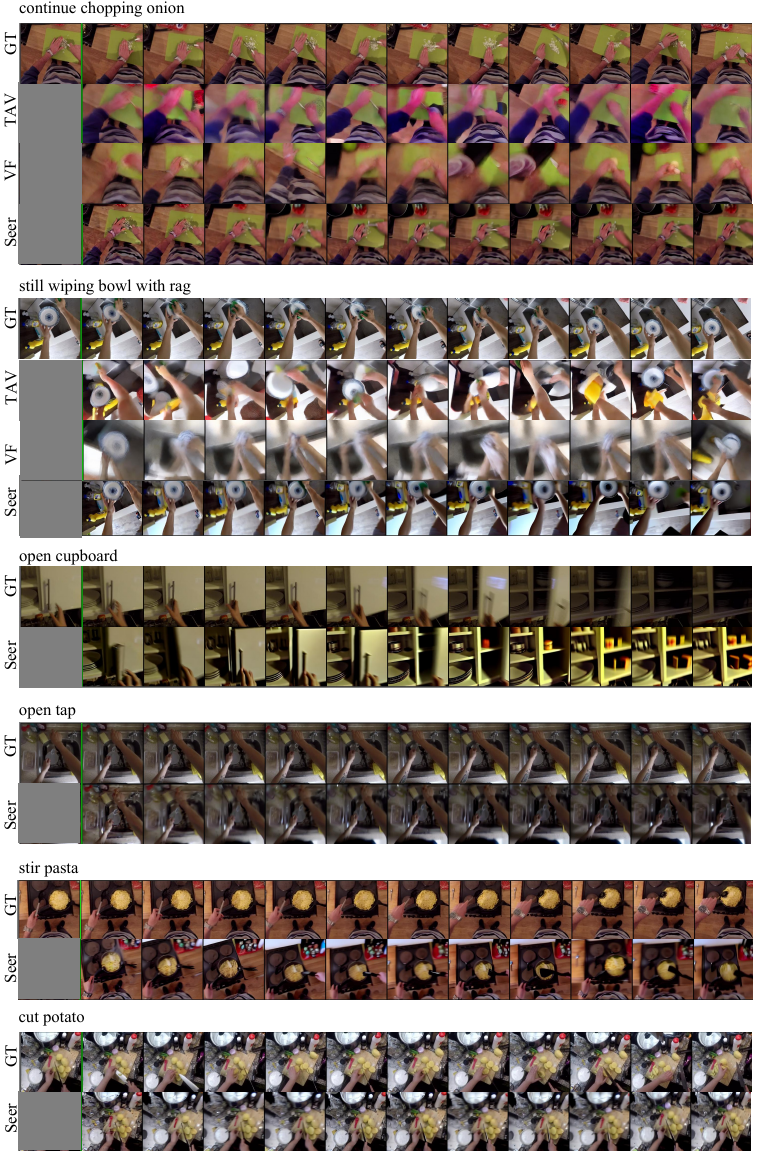}
\vspace{-8pt}
\caption{Text-conditioned video prediction (conditioned on first frame)
on Epic-Kitchens-100. TAV refers to Tune-A-Video, VF indicates VideoFusion.}
\label{fig:epicpred}
\end{figure*}

To evaluate the quality of video predictions according to human preferences, we conducted a human evaluation with 99 video clips on the validation set of the Something-Something V2 dataset (SSv2), the evaluation process involved 54 anonymous evaluators. To eliminate biases towards specific baselines, we randomly selected 20 questions for each evaluator. Each single-choice question consisted of a ground-truth video as a reference, a manually modified text instruction, and two video prediction results generated by Seer and another baseline method. The evaluators were required to choose the video clip that is more consistent with the text instruction and has higher fidelity from the two options.
To ensure the clarity of the questions, we provided an example to explain the options in each questionnaire. Moreover, we recommended that evaluators prioritize video predictions with strong text-based motions as their first preference and the fidelity of the generated video as their second preference. For reference, Figure~\ref{fig:humanevalexp} provides a screenshot of an example questionnaire.

In total, we collected 342 responses for the Seer vs. TATS comparison, 363 responses for the Seer vs. Tune-A-Video comparison, and 357 responses for the Seer vs. MCVD comparison. And the results in the main paper Figure 7 are calculated based on the collected questionnaires.
\clearpage

\end{document}